\documentclass{article}
\usepackage[usenames]{color}
\definecolor{plum}  {rgb}{.4,0,.4}
\definecolor{forest}  {rgb}{0,.6,0}
\definecolor{midnight}  {rgb}{0,0,.8}
\definecolor{brick}  {rgb}{.8,0,0}

\usepackage[pdftex,plainpages=false,pdfpagelabels,colorlinks=true,urlcolor=midnight,linkcolor=brick,citecolor=midnight]{hyperref} 
\usepackage{float,subfig}
\usepackage{amssymb,amsmath,enumerate,cite,url,bm,color,graphicx,bbm}
\usepackage{algorithm,algorithmic} 

\newcommand{\zeros}{\bm{0}}

\def\htheta{{\widehat{\theta}}}
\def\ttheta{{\widetilde{\theta}}}

\def\THETA{\boldsymbol{\theta}}

\def\argmin{\mathop{\!\arg \min}}

\def\deq{{\triangleq}}
\def\grad{\nabla}
\def\diam{D_{\rm max}}

\def\reals{\mathbb{R}}
\def\btheta{\boldsymbol{\theta}}

\def\bhtheta{\widehat{\btheta}}

\newcommand{\wh}[1]{\widehat{#1}}
\newcommand{\wt}[1]{\widetilde{#1}}

\newcommand{\ave}[1]{\langle #1 \rangle}

\def\cX{{\mathsf X}}

\def\Int{\text{Int }}
\def\ones{\boldsymbol{1}}

\newtheorem{theorem}{{\bf Theorem}}

\newtheorem{definition}[theorem]{{\bf Definition}}
\newtheorem{lemma}[theorem]{{\bf Lemma}}
\newtheorem{corollary}[theorem]{{\bf Corollary}}

\newcommand{\ie}[0]{\emph{i.e.,} }

\newcommand{\eg}[0]{\emph{e.g.,} }

\newcommand{\bseq}{\begin{subequations}\begin{align}}
\newcommand{\eseq}{\end{align}\end{subequations}}

\title{Online Convex Optimization in\\ Dynamic Environments}

\author{Eric~C.~Hall\thanks{E. C. Hall is with the Department of Electrical and Computer
    Engineering, Duke University, Durham, NC, 27708, USA. e-mail:
    ech11@duke.edu},~and~Rebecca~M.~Willet\thanks{R. M. Willett is with the Department of
    Electrical and Computer Engineering, University of
    Wisconsin-Madison, Madison, WI 53706, USA. e-mail:
    willett@discovery.wisc.edu. Portions of this paper appeared in an ICML paper
    \cite{dynamicMirrorDescent}, which proposed the DMD method and the
    fixed-share approach to choosing among dynamical models. The
    current paper builds substantially upon that paper. We (a) more fully discuss
    the dynamical model $\Phi$ to changing with time, opening the door
    to new analyses, including those related to autoregressive moving
    average (ARMA) models and self-exciting point processes, (b)
    consider parametric families of dynamical models and a covering
    technique for learning the best dynamical model, (c) develop
    particularly efficient methods and bounds for additive dynamical
    models with exponential family losses, and (d) present several new
    experimental results. We gratefully acknowledge the support of the awards AFSOR 14-AFOSR-1103
and NSF CCF-1418976.}}
\date{}
  
\begin{document}
\maketitle

\begin{abstract}
  High-velocity streams of high-dimensional data pose significant
  ``big data'' analysis challenges across a range of applications and
  settings. Online learning and online convex programming play a
  significant role in the rapid recovery of important or anomalous
  information from these large datastreams. While recent advances in online
  learning have led to novel and rapidly converging algorithms, these
  methods are unable to adapt to nonstationary environments arising in
  real-world problems. This paper describes a {\em dynamic mirror
    descent} framework which addresses this challenge, yielding low
  theoretical regret bounds and accurate, adaptive, and
  computationally efficient algorithms which are applicable to broad
  classes of problems. The methods are capable of learning and
  adapting to an underlying and possibly time-varying dynamical
  model. Empirical results in the context of dynamic texture analysis, solar flare detection,
  sequential compressed sensing of a dynamic scene, traffic surveillance,tracking
  self-exciting point processes and network behavior in the Enron email corpus support the core
  theoretical findings.
\end{abstract}


\section{Introduction}
Modern sensors are collecting very high-dimensional data at
unprecedented rates, often from platforms with limited processing
power. These large datasets allow scientists and analysts to consider
richer physical models with larger numbers of variables, and thereby
have the potential to provide new insights into the underlying complex
phenomena.  For example, the Large Hadron Collider (LHC) at CERN
``generates so much data that scientists must discard the overwhelming
majority of it -- hoping hard they've not thrown away anything
useful.'' \cite{LHC} Typical NASA missions collect hundreds of
terabytes of data every hour \cite{JPLBigData}: the Solar Data
Observatory generates 1.5 terabytes of data daily \cite{SDO}, and the upcoming
Square Kilometer Array (SKA, \cite{SKA}) is projected to generate an
exabyte of data daily, ``more than twice the information sent around
the internet on a daily basis and 100 times more information than the
LHC produces'' \cite{WiredSKA}.  In these and a variety of other science and engineering settings,
there is a pressing need to recover {\em relevant or anomalous}
information {\em accurately and efficiently} from a high-dimensional,
high-velocity data stream.

Rigorous analysis of such data poses major issues, however.  First, we
are faced with the notorious ``curse of dimensionality'', which states
that the number of observations required for accurate inference in a
stationary environment grows exponentially with the dimensionality of
each observation. This requirement is often unsatisfied even in
so-called ``big data'' settings, as the underlying environment varies
over time in many applications. Furthermore, any viable method for
processing massive data must be able to scale well to high data
dimensions with limited memory and computational resources.  Finally,
in a variety of large-scale streaming data problems, ranging from
motion imagery formation to network analysis, the underlying
environment is dynamic yet predictable, but many general-purpose and
computationally efficient methods for processing streaming data lack a
principaled mechanism for incorporating dynamical models.  Thus a
fundamental mathematical and statistical challenge is accurate and
efficient tracking of dynamic environments with high-dimensional
streaming data.

Classical stochastic gradient descent methods, including the least mean squares
(LMS) or recursive least squares (RLS) algorithms do not have a
natural mechanism for incorporating dynamics. Classical stochastic
filtering methods such as Kalman or particle filters or Bayesian
updates \cite{BaiCri09} readily exploit dynamical models for effective
prediction and tracking performance.  However, these methods are
also limited in their applicability because (a) they typically assume
an accurate, fully known dynamical model and (b) they rely on strong
assumptions regarding a generative model of the observations. Some
techniques have been proposed to learn the dynamics
\cite{XieSoh94,TheSha96}, but the underlying model still places heavy
restrictions on the nature of the data. Performance analysis of these
methods usually does not address the impact of ``model mismatch'',
where the generative models are incorrectly specified.

A contrasting class of prediction methods, receiving widespread recent
attention within the machine learning community, is based on an
``individual sequence'' or ``universal prediction'' \cite{MerFed98}
perspective; these strive to perform provably well on any individual
observation sequence without assuming a generative model of the data. 

{\em Online convex programming} provides a variety of tools for
sequential universal prediction
\cite{NemYud83,BecTeb03,Zin03,CesLug06}. Here, a Forecaster measures
its predictive performance according to a convex loss function, and
with each new observation it computes the negative gradient of the
loss and shifts its prediction in that direction. Stochastic gradient
descent methods stem from similar principles and have been studied for
decades, but recent technical breakthroughs allow these approaches to
be understood without strong stochastic assumptions on the data, even
in adversarial settings, leading to more efficient and rapidly
converging algorithms in many settings.

This paper describes a novel framework for prediction in the
individual sequence setting which incorporates dynamical models --
effectively a novel combination of state updating from stochastic
filter theory and online convex optimization from universal
prediction.  We establish tracking regret bounds for our proposed
algorithm, {\em Dynamic Mirror Descent (DMD)}, which characterize how
well we perform relative to some alternative approach (\eg a
computationally intractable batch algorithm) operating on the same
data to generate its own predictions, called a ``comparator
sequence.''  Our novel regret bounds scale with the deviation of this
comparator sequence from a dynamical model.  These bounds simplify to
previously shown bounds when there are no dynamics. 
In
addition, we describe methods based on DMD for adapting to the best
dynamical model from either a finite or parametric class of candidate
models. In these settings, we establish tracking regret bounds which
scale with the deviation of a comparator sequence from the {\em best
  sequence} of dynamical models.

While our methods and theory apply in a broad range of settings, we
are particularly interested in the setting where the dimensionality of
the parameter to be estimated is very high.  In this regime, the
incorporation of both dynamical models and sparsity regularization
plays a key role. With this in mind, we focus on a class of methods
which incorporate regularization as well as dynamical modeling. The
role of regularization, particularly sparsity regularization, is
increasingly well understood in batch settings and has resulted in
significant gains in ill-posed and data-starved settings
\cite{BanGhaAsp08,RavWaiLaf10,CS:noiseEC,BelNiy03}. More recent work
has examined the role of sparsity in online methods such as recursive
least squares (RLS) algorithms, but do not account for dynamic
environments \cite{sparseRLS}.

\subsection{Organization of paper and main contributions}
The remainder of this paper is structured as follows. In
Section~\ref{sec:prob}, we formulate the problem and introduce
notation used throughout the paper, and Section~\ref{sec:oco} introduces the {\em
  Dynamic Mirror Descent (DMD)} method, and gives brief comparison to existing methods. along with novel tracking regret bounds.
This section also describes the application of
data-dependent dynamical models and their connection to recent work on
online learning with predictable sequences. DMD uses only a single
series of dynamical models, but we can use it to choose among a family
of candidate dynamical models. This is described for finite families
in Section~\ref{sec:PWEA} using a fixed share algorithm, and for
parametric families in
Section~\ref{sec:para}. Section~\ref{sec:results} shows experimental
results of our methods in a variety of contexts ranging from imaging
to self-exciting point processes. Finally, Section~\ref{sec:conc} makes
concluding remarks while proofs are relegated to
Section~\ref{sec:proofs}.

\section{Problem formulation}
\label{sec:prob}

The problem of sequential prediction is posed as an iterative game
between a Forecaster and the Environment.  At every time point, $t$,
the Forecaster generates a prediction $\htheta_t$ from a bounded, closed,
convex set $\Theta \subset \reals^d$.  After the Forecaster makes a
prediction, the Environment reveals the loss function $\ell_t(\cdot)$
where $\ell_t$ is a convex function which maps the space $\Theta$ to
the real number line. We will assume that the loss function is the
composition of a convex function $f_t: \Theta \rightarrow \reals$ from
the Environment and a convex regularization function $r: \Theta
\rightarrow \reals$ which does not change over time. Frequently the
loss function, $f_t$ will measure the accuracy of a prediction
compared to some new data point $x_t \in \cX$ where $\cX$ is the
domain of possible observations. The regularization function promotes
low-dimensional structure (such as sparsity) within the
predictions. We additionally assume that we can compute a subgradient
of $\ell_t$ or $f_t$ at any point $\theta \in \Theta$, which we denote $\grad
\ell_t$ and $\grad f_t$. Thus
the Forecaster incurs the loss
$\ell_t(\htheta_t)=f_t(\htheta_t)+r(\htheta_t)$.

The goal of the Forecaster is to create a sequence of predictions
$\htheta_1,\htheta_2,\ldots,\htheta_T$ that has a low cumulative loss
$\sum_t ^T\ell_t(\htheta_t)$. Because the loss functions are being
revealed sequentially, the prediction at each time can only be a
function of all previously revealed losses to ensure causality.  Thus,
the task facing the Forecaster is to create a new prediction,
$\htheta_{t+1}$, based on the previous prediction and the new loss
function $\ell_t(\cdot)$, with the goal of minimizing loss at the next
time step.  We characterize the efficacy of $\wh{\btheta}_T \deq
(\htheta_{1}, \htheta_{2},\ldots,\htheta_T) \in \Theta^T$ relative to
a comparator sequence $\btheta_T\deq
(\theta_1,\theta_2,\ldots,\theta_T) \in \Theta^T$ using a concept
called {\em regret}, which measures the difference of the total
accumulated loss of the Forecaster with the total accumulated loss of
the comparator:
\begin{definition}[Regret] \label{defn:regret} The {\em regret} of
  $\bhtheta_T$ with respect to a comparator $\btheta_T \in \Theta^T$
  is
$$  R_T(\btheta_T) \deq  \sum_{t=1}^{T} \ell_{t}(\htheta_{t}) - \sum_{t=1}^{T}\ell_{t}(\theta_{t}).$$
\end{definition}

Notice that this definition of regret is very general and simply
measures the performance of an algorithm versus an arbitrary sequence
$\btheta_T$. We are particularly interested in comparators
  which correspond to the output of a {\em batch} algorithm (with
  access to all the data simultaneously) that is
too computationally complex or memory-intensive for practical big data
analysis problems. In this sense,
regret encapsulates how much one regrets working in an online setting
as opposed to a batch setting with full knowledge of past and future
observations.

Much of the online learning literature is focused on
  algorithms with guaranteed sublinear regret (\eg $R_T(\btheta_T) =
  O(\sqrt{T})$) in the special case where the comparator $\btheta_T$
  is constrained so that $\theta_1 = \theta_2 = \ldots =
  \theta_T$. Unfortunately, this is a highly unrealistic constraint in
most practical streaming big data settings. The parameters $\theta_t$
could correspond to frames in a video or the weights of edges in a
dynamic network and by nature are highly variable.

This paper focuses more generally on arbitrary comparator
  sequences $\btheta_T$ and shows how the regret scales as a function
  of the temporal variability in that comparator. This idea is
  typically referred to as ``tracking'' or ``shifting'' regret
  \cite{HerWar01,CesGaiLugSto12}, which is closely-related to ``adaptive'' regret \cite{LitWar94,HazSes09}. Existing methods
  guarantee sublinear regret for all $\btheta_T$ which do not vary at
  all over time, or which change only at a few discrete points in
  time, or which only vary extremely slowly over time; generally,
  these regret bounds depend linearly on a {\em variation} term of the form
$$V(\btheta_T) \deq \sum_{t=1}^{T-1} \|\theta_{t+1} - \theta_t\|;$$
sublinear regret is only possible for comparator sequences for which
this variation is small.  {\em In contrast, the proposed methods in this
paper guarantee sublinear regret for different classes of
$\btheta_T$ that allow quite large temporal variability.}

\section{Dynamic Mirror Descent}\label{sec:oco}
To begin, we propose a simple modification to the ``mirror
  descent'' online learning paradigm. Specifcally, we incorporate a dynamical
  model $\Phi_t$ at each time $t$; this approach is called {\em
    dynamic mirror descent (DMD)}. For now, we assume $\Phi_t$ is
  fixed and known. For example,
\begin{itemize}
\item Streaming observations correspond to (potentially
  distorted) frames in a video sequence, $\theta_t$ corresponds to
  a set of wavelet coefficients for that frame, and $\Phi_t$
  corresponds to a video motion model at time $t$.
\item Streaming observations correspond to interactions within a
 social network, $\theta_t$ corresponds the likelihood of each
pair of  people interacting at time $t$, and $\Phi_t$ captures diurnal
patterns and social network evolution models. 
\item Streaming observations correspond to the price of stocks,
  $\theta_t$ parameterizes a probability distribution governing stock
  prices, and $\Phi_t$ captures underlying autoregressive behavior and
  side information derived from other financial instruments.
\end{itemize}
In all of these settings, it is possible to posit a dynamical model
$\Phi_t$ based on prior knowledge of the application, akin to
developing a state space model for stochastic filters.

\begin{algorithm}
\caption{Dynamic mirror descent (DMD) with known dynamics}
\label{alg:dmd}
\begin{algorithmic} 
\STATE Given non-increasing sequence of step sizes $\eta_t > 0$
\STATE Initialize $\wh{\theta}_{1} \in \Theta$.
\FOR {$t=1,\ldots,T$}
\STATE Observe $x_t$ and incur loss $\ell_t(\wh{\theta}_t)$
\STATE Environment produces dynamical model $\Phi_t$
\STATE Set \begin{subequations}
\label{eq:dmd}
\begin{align}
\ttheta_{t+1} &= \argmin_{\theta \in \Theta} \eta_t
  \ave{\grad f_t(\htheta_t),\theta} + \eta_t r(\theta) + D(\theta \|
  \htheta_t)\label{eq:dmd1}\\
  \htheta_{t+1} &= \Phi_{t}( \ttheta_{t+1}) \label{eq:dmd2}
\end{align}
\end{subequations}
\ENDFOR
\end{algorithmic}
\end{algorithm}

The DMD algorithm is presented in
Algorithm~\ref{alg:dmd}. In this algorithm $\grad f_t(\theta)$ denotes an arbitrary subgradient of
$f_t$ at $\theta$, $D(\theta \| \htheta_t)$ is the {\em Bregman
  divergence} \cite{Bre67,CenZen97} which measures the distance between $\theta$ and $\htheta$,
and $\eta_t > 0$ is a step size parameter. 

We may compare the DMD approach with classical online
  learning methods \cite{NemYud83,LitWar94,Zin03,BecTeb03} and more
  recent reguarlized formulations
  \cite{COMD,xiao,Lang09}. For instance,
 mirror descent
  (MD, 
  \cite{NemYud83,BecTeb03}) sets \begin{align}
\htheta_{t+1} = \argmin_{\theta \in \Theta} \eta_t \ave{\grad
  \ell_t(\htheta_t),\theta} + D(\theta \| \htheta_t),
\label{eq:md}
\end{align}
and 
Composite Objective
Mirror Descent (COMID\footnote{The COMID formulation is helpful when the regularization function
$r(\theta)$ promotes sparsity in $\theta$, and helps ensure that $\htheta_t$ is indeed sparse.}, \cite{COMD}) sets
\begin{align}
  \htheta_{t+1} = \argmin_{\theta \in \Theta} \eta_t
  \ave{\grad f_t(\htheta_t),\theta} + \eta_t r(\theta) + D(\theta \|
  \htheta_t).
\label{eq:comd}
\end{align}
Specifically, note that if $\Phi_t$ is the identity operator for all
$t$, so that $\Phi_t( \theta) \equiv \theta$ for all $\theta$ and $t$,
then DMD corresponds exactly to COMID.

Methods like MD and COMID have sublinear regret bounds only for
comparators with small $V(\btheta_T)$, such as when $\theta_t =
\theta$ for some $\theta \in \Theta$ and all $t$.  In contrast, our
method is an algorithm which incorporates a dynamical model, denoted
$\Phi_t: \Theta \mapsto \Theta$, and admits a tracking regret bound of
the form $O(\sqrt{T}[1 + \sum_{t=1}^{T-1} \| \theta_{t+1} - \Phi_t(
\theta_t)\|])$ (shown in the next section).

By including $\Phi_{t}$ in the process, we effectively search for a predictor which (a)
attempts to minimize the loss and (b) which adheres to the dynamical model $\Phi_t$.
The DMD approach effectively includes dynamics into the COMID
framework.\footnote{Rather than considering COMID, we might have used
  other online optimization algorithms, such as the Regularized Dual
  Averaging (RDA) method \cite{xiao}, which has been shown to achieve
  similar performance with more regularized solutions.  However, to
  the best of our knowledge, no tracking or shifting regret bounds
  have been derived for dual averaging methods (regularized or
  otherwise). Recent results on the equivalence of COMID and RDA
  \cite{McMahan11} suggest that the bounds derived here might also
  hold for a variant of RDA, but proving this remains an open
  problem.} 

The important feature of Alg. \ref{alg:dmd} is Equation
  \ref{eq:dmd2}, where the predetermined function $\Phi_t$ is
  incorporated in the learning process.  The main intuition is that
  instead of believing the next loss function is well approximated by
  all the previous loss functions as in MD and COMID, we are instead
  assuming the loss functions will be well predicted by the trajectory
  encoded by the functions $\Phi_t$.  This results in changing the
  class of comparators which lead to sublinear regret bounds.  MD and
  COMID have sublinear regret when the comparator $\btheta_T$ is
  static or changes very slowly.  However, as we will show, DMD yields
  sublinear regret when the comparator evolves according to the
  functions $\Phi_t$ or only deviates from these functions by a small
  amount, meaning $\sum_{t=1}^T\|\theta_{t+1}-\Phi_t(\theta_t)\|$ is small. Thus, while the class of $\btheta_T$ that has low regret may
  be the same size as with MD or COMID, the classes contain very
  different comparator sequences. In later sections we will
  address how to learn the sequence $\Phi_t$ from the data.
  Additionally, we make no assumption about whether the {\em data}
  actually follow these dynamics, but instead we derive a regret bound
  which scales with how well the {\em comparator} evolves according to
  these functions.

\subsection{Tracking regret of DMD}
\label{sec:DMD}

Our main result uses the following assumptions:
\begin{itemize}
\item Let $\psi:\Theta
\rightarrow \reals$ denote a continuously differentiable function that
is $\sigma$-strongly convex for some parameter $\sigma>0$ and some
norm $\|\cdot\|$
$$
\psi(\theta_1) \geq \psi(\theta_2) + \langle \grad
\psi(\theta_2),\theta_1 - \theta_2 \rangle + \frac{\sigma}{2}
\|\theta_1 - \theta_2\|^2 \label{eq:psi_strong} 
$$
\item The Bregman Divergence used in our algorithm is defined as $D(\theta_1\| \theta_2) 
\deq
\psi(\theta_1) - \psi(\theta_2) -
\ave{\nabla\psi(\theta_2),\theta_1-\theta_2}$. Because $\psi$ is $\sigma$-strongly convex we have 
\begin{align}
D(\theta_1\|\theta_2) \geq \frac{\sigma}{2}\|\theta_1-\theta_2\|^2 \label{eq:convBreg}
\end{align}
Additionally, the definition of the Bregman Divergence implies the following relationship: for all $\theta_1,\theta_2,\theta_3 \in \Theta$
\begin{align}
D(\theta_1\|\theta_2)=& D(\theta_3\|\theta_2) +
D(\theta_1\|\theta_3) \nonumber \\
& +
\ave{\grad\psi(\theta_2)-\grad\psi(\theta_3),\theta_3-\theta_1}.
\label{eq:triBreg} 
\end{align}
\item For all $t=1,\ldots,T$ the functions $\ell_t$ and $\psi$ are Lipschitz with constants $G$ and $M$ respectively, such that $\|\grad \ell_t (\theta)\|_* \leq G$ and $\|\grad\psi(\theta)\|_* \leq M$ for all $\theta\in \Theta$.  The function $\|\cdot\|_*$ used in these assumptions is the dual to the norm that $\psi$ is strongly convex with respect to.
\item There exists a constant $D_{\max}$ such that $D(\theta_1\|\theta_2) \leq D_{\max}$ for all $\theta_1,\theta_2 \in \Theta$.
\item For all $t=1,\ldots,T$, the transformation $\Phi_t$ has a
  maximum distortion factor $\Delta_{\Phi}$ such that
  $D(\Phi_t(\theta_1)\|\Phi_t(\theta_2)) - D(\theta_1\|\theta_2) \leq
  \Delta_{\Phi}$ for all $\theta_1,\theta_2 \in \Theta$.  When $\Delta_\Phi \leq 0$ for all $t$, we say that
  $\Phi_t$ satisfies the contractive property.
\end{itemize}

\begin{theorem}[Tracking regret of dynamic mirror descent]
  \label{thm:main} Let $\Phi_t$ be a dynamical model such that
  $\Delta_{\Phi} \leq 0$ for $t = 1, 2, \ldots,T$ with respect to
  the Bregman divergence used in \ref{eq:dmd}. Let the sequence
  $\bhtheta_T$ be generated using Alg.~\ref{alg:dmd} using a
  non-increasing series $\eta_{t+1} \leq \eta_{t}$, with a convex,
    Lipschitz function $\ell_t$ on a closed, convex, bounded set $\Theta$, and
    let $\btheta_T$ be an arbitrary sequence in $\Theta^T$. Then
\begin{gather*}
R_T(\btheta_T) \leq \frac{\diam}{\eta_{T+1}} + \frac{2M}{\eta_T} V_{\Phi}(\btheta_T) +
\frac{G^2}{2\sigma} \sum_{t=1}^{T} \eta_t \\
\mbox{with }
V_{\Phi}(\btheta_T) \deq \sum_{t=1}^{T-1} \|\theta_{t+1} - \Phi_t(\theta_t)\| 
\end{gather*}
where $V_{\Phi}(\btheta_T)$
measures variations or deviations of the comparator sequence $\btheta_T$ from the sequence of
dynamical models $\Phi_1, \Phi_2, \ldots, \Phi_T$. If $\eta_t \propto \frac{1}{\sqrt{t}}$ or $\eta_t \propto \frac{1}{\sqrt{T}}$, then for some $C>0$ independent of $T$,
\begin{align*}
R_T(\btheta_T) \leq C\sqrt{T}\left(1+V_{\Phi}(\btheta_T)\right)
\end{align*}
\end{theorem}

This bound scales with the comparator sequence's deviation from the
sequence of dynamical models $\{\Phi_t\}_{t>0}$ -- a stark contrast to
previous tracking regret bounds which are only sublinear for
comparators which change slowly with time or at a small number of
distinct time instances.  Note that when $\Phi_t$ corresponds to an
identity operator, the bound in Theorem~\ref{thm:main} corresponds to
existing tracking or shifting regret bounds
\cite{CesLug06,CesGaiLugSto12}.

It is intuitively satisfying that this measure of variation,
$V_{\Phi}(\btheta_T)$, appears in the tracking regret bound.  First,
if the comparator sequence evolves approximately like $\theta_{t+1}=\Phi_t(\theta_t)$, this variation term
will be very small, leading to low regret.  This fact holds
whether $\Phi_t$ is part of the generative model for the observations
or not. Secondly, we can get a dynamic analog of static regret, where we
enforce $V_\Phi(\btheta_T) = 0$.  This is equivalent to saying that
the batch comparator is fitting the best single trajectory using
$\Phi_t$ instead of the best single point.  Using this, we would
recover a bound analogous to a static regret bound in a stationary
setting.

The condition that $\Delta_{\Phi} \leq 0$ is similar to requiring that
$\Phi_t$ be a contractive mapping.  This restriction is important;
without it, any poor prediction made at one time step could be
exacerbated by repeated application of the dynamics.  For instance,
linear dynamic models with all eigenvalues
less than or equal to unity satisfy this condition with respect to the
squared $\ell_2$ Bregman Divergence, similar in spirit to restrictions
made in more classical adaptive filtering work such as
\cite{Haykin02}. Notice also that if $\Phi_t(\theta)=\theta$ in for all $t$, then Theorem
\ref{thm:main} gives a novel, previously unknown tracking regret bound
for COMID.  
\subsection{Data-dependent dynamics}
An interesting example of dynamical models is the class of
data-dependent dynamical models.  In this regime the state of the
system at a given time is not only a function of the previous state,
but also the actual observations.  One key example of this scenario
arises in self-exciting point processes, where the state of the system
is directly related to the previous observations. Our
algorithm can account for such models since the function
$\Phi_t(\theta)$ is time varying, and therefore can implicitly depend
on all data up to time $t$,
i.e. $\Phi_t(\theta)=\Phi_t(\theta,x_1,x_2,\ldots,x_t)$.  Our regret
bounds therefore scale with how well the comparator series matches
these data dependent dynamics:
\begin{align*}
R_T(\btheta_T) \leq C\left(\sqrt{T}\left[1+\sum_{t=1}^{T-1}\|\theta_{t+1}- \Phi_t(\theta_t,x_1,\ldots,x_t)\|\right]\right).
\end{align*}
Notice now that the data plays a part in the regret bounds, whereas
before we only measured the variation of the comparator.
Data-dependent regret bounds are not new. Concurrent related work
considers online algorithms where the data sequence is described by a
``predictable process" \cite{Rak12}. The basic idea of that paper is
that if one has a sequence of functions $M_t$ which predict $x_{t}$ based
on $x_1,x_2,\ldots,x_{t-1}$, then the output of a standard online
optimization routine should be combined with the predictor generated
by $M_t$ to yield tighter regret bounds that scale with $(\sum_t \|x_t
- M_t(x_1,\ldots,x_{t-1})\|^2)^{1/2}$.  However, \cite{Rak12} only
works with static regret (\ie regret with respect to a static
comparator) and their regret has a variation term that expresses the
deviation of the {\em input data} from the underlying process. In
contrast, our tracking regret bounds scale with the deviation of a
{\em comparator sequence} from a prediction model. 

\section{Prediction with a finite family of  dynamical models}
\label{sec:PWEA}
DMD in the previous section uses a single sequence of dynamical
models. In practice, however, we may not know the best dynamical model
to use, or the best model may change over time in nonstationary
environments.  To address this challenge, we assume a finite set of
candidate dynamical models $\{\Phi_{1,t}, \Phi_{2,t}, \ldots
,\Phi_{N,t}\}$ at every time $t$, and describe a procedure which uses
this collection to adapt to nonstationarities in the environment.  In
particular we establish tracking regret bounds which scale not with
the deviation of a comparator from a single dynamical model, but with
how it deviates from a {\em series of different dynamical models on
  different time intervals} with at most $m$ switches.  These switches
define $m+1$ different time segments $[t_i,t_{i+1}-1]$ with time
points $1= t_1 < \cdots < t_{m+2} = T$.  We can bound the regret associated with the best dynamical model on each time segment and then bound the overall regret using a Prediction with Experts Advice algorithm.

Our {\em dynamic fixed share} (DFS) estimate
is presented in Algorithm~\ref{alg:dfs}.
Let $\htheta_{i,t}$ denote the output of Equation \ref{eq:dmd2} at time $t$ using
dynamical models $\Phi_{i,1}, \Phi_{i,2}, \ldots, \Phi_{i,t}$; we
choose $\htheta_t$ by using the Fixed Share forecaster on these
outputs.\footnote{There are many algorithms from the Prediction with
  Expert Advice literature which can be used to form a single
  prediction from the predictions created by the set of dynamical
  models.  We use the Fixed Share algorithm \cite{HerWar98} as a means
  to combine estimates with different dynamics; however, other methods
  could be used with various tradeoffs.  One of the primary drawbacks
  of the Fixed Share algorithm is that an upper bound on the number of
  switches $m$ must be known a priori.  However, this method has a
  simple implementation and tracking regret bounds.  One common
  alternative to Fixed Share allows the switching parameter ($\lambda$
  in Alg.~\ref{alg:dfs}) to decrease to zero as the algorithm runs
  \cite{AdamRooij08,Adam12}.  This has the benefit of not requiring
  knowledge about the number of switches, but comes at the price of
  higher regret. 
  Alternative expert advice algorithms exist which decrease the regret
  but increase the computational complexity.  For a thorough treatment
  of existing methods see \cite{Gyo12}.}  In Fixed Share, each expert (here,
each sequence of candidate dynamical models) is assigned a weight that
is inversely proportional to its cumulative loss at that point yet
with some weight shared amongst all the experts.  Here there is $\lambda$ amount of weight divided evenly amongst experts so that an expert
with previously high loss quickly regain enough weight to become the leader
\cite{HerWar98,CesGaiLugSto12}.
In this update, $\lambda \in (0,1)$ is how much of the weight is shared amongst the experts. Notice each expert ``donates" a $\lambda$ fraction of it's weight which is then replaced by $\lambda/N$.  For experts which large weights this will cause some weight to be lost, but ensures that a minimum weight of $\lambda/N$ is attained for each expert. Sharing weight allows fast switching between experts.
\begin{algorithm}
\caption{Dynamic fixed share (DFS)}
\label{alg:dfs}
\begin{algorithmic} 
\STATE Given decreasing sequence of step sizes $\eta_t > 0$ and
$\eta_r > 0$
\STATE Initialize $\wh{\theta}_{1} \in \Theta$,
$\wh{\theta}_{i,1} \in \Theta$ and $w_{i,1} = \frac{1}{N}$ for
$i=1,\ldots,N$, $\lambda \in (0,1)$, and
$\eta_t,\eta_r > 0$.
\FOR {$t=1,\ldots,T$}
\STATE Observe $x_t$ and incur loss $\ell_t(\wh{\theta}_t)$
\STATE Receive dynamical model $\Phi_{i,t}$ for $i = 1,\ldots,N$
\FOR {$i = 1,\ldots,N$}
\STATE Set 
\begin{align*}
\wt{w}_{i,t+1}&=\frac{w_{i,t}\exp\left(-\eta_r \ell_t\left(\htheta_{i,t}\right)\right)}{\displaystyle\sum_{j=1}^N w_{j,t}\exp\left(-\eta_r \ell_t\left(\htheta_{j,t}\right)\right)}\\
w_{i,t+1}&=\frac{\lambda}{N}+ (1-\lambda)\wt{w}_{i,t+1}\\
\ttheta_{i,t+1} &= \argmin_{\theta \in \Theta} \eta_t
  \ave{\grad f_t(\htheta_{i,t}),\theta} + \eta_t r(\theta) + D(\theta \|
  \htheta_{i,t})\\
  \htheta_{i,t+1} &= \Phi_{i,t}( \ttheta_{i,t+1}) 
\end{align*}
\ENDFOR
\STATE Set $$
\htheta_{t+1}=\displaystyle\sum_{i=1}^N w_{i,t+1}\htheta_{i,t+1} 
$$
\ENDFOR
\end{algorithmic}
\end{algorithm}
\begin{theorem}[Tracking regret of DFS algorithm]
\label{thm:dfs}
Assume all the candidate dynamic sequences are contractive such that $\Delta_{\Phi}\leq0$ for $\Phi_{i,t}$ for all $t=1,..,T$ and $i=1,...,N$ with respect to the Bregman divergence in Alg \ref{alg:dmd}.  Then for some $C>0$, the dynamic fixed share algorithm in Algorithm~\ref{alg:dfs} with 
parameter $\lambda$ set equal to $\frac{m}{T-1}$,
$\eta_r=\sqrt{\frac{8\left((m+1)\log(N)+m\log(T)+1\right)}{T}}$ and
    $\eta_t \propto 1/\sqrt{t}$ or $\eta_t \propto 1/\sqrt{T}$ with a convex, Lipschitz function $\ell_t$
    on a closed, bounded, convex set $\Theta$, has tracking regret 
\begin{align*}
&R_T(\btheta_T) = \sum_{t=1}^{T} \ell_t(\htheta_t) - 
  \sum_{t=1}^{T} \ell_t(\theta_t) \\
&\leq C\left(\sqrt{T}\left(\sqrt{(m+1)\log N + m \log T}+  V^{(m+1)}(\btheta_T)\right) \right),
\end{align*}
where
\begin{align*}V&^{(m+1)}(\btheta_T) \deq
\min_{t_2, \ldots, t_{m+1}}
\sum_{k=1}^{m+1} \min_{i_k \in \{1,\ldots,N\}} \sum_{t=t_k}^{t_{k+1}-1} \|
\theta_{t+1}-\Phi_{i_k,t}(\theta_t) \|
\end{align*}

measures the deviation of the sequence
$\btheta_T$ from the best sequence of dynamical models with at most
$m$ switches (where $m$ does not depend on $T$). 
\end{theorem}

The choice of $m$ is important, as low values of $m$ will have low regret but for a smaller class of comparators, comparators with a small number of switches. Oppositely, larger values of $m$ will have low regret for a larger class of comparators, but there is an overhead to be paid in the constant terms.
Note that the family of comparator sequences $\btheta_T$ for which
$R_T(\btheta_T)$ scales sublinearly in $T$ is {\em significantly}
larger than the set of comparators yielding sublinear regret for MD. This is because $
V^{(m+1)}(\btheta_T) \leq V_{\Phi_{i,t}}(\btheta_T)$ for any fixed $i \in
\{1,\ldots,N\}$, thus this approach yields a lower variation term than
using a fixed dynamical model.  However, we incur some loss by not
knowing the optimal number of switches $m$ or when the optimal switching times are.

\section{Parametric dynamical models}
\label{sec:para}
Rather than having a finite family of dynamical models, as we did in
Section~\ref{sec:PWEA}, we may consider a parametric family of
dynamical models, where the parameter $\alpha \in \reals^n$ of
$\Phi_t$ is allowed to vary across a closed, bounded, convex domain,
denoted $\mathcal{A}$. In other words, we consider $\Phi_t: \Theta
\times \mathcal{A} \mapsto \Theta$. In this context we would like to
{\em jointly} predict both $\alpha$ and $\theta$. 

We consider two approaches. First, in Section~\ref{sec:grid} we consider
tracking only a finite subset of the possible
model parameters, in a manner similar to when we had a finite collection of 
possible dynamical models, which provide a ``covering'' of the parameter space. In this case, the overall regret and computational complexity both depend on the resolution of the covering set. Second, in Section~\ref{sec:additive}, we consider a special family of additive dynamical models; in this setting, we can efficiently learn the optimal dynamics. 

\subsection{Covering the set of dynamical models}\label{sec:grid}

In this section we show that by tracking a subset which appropriately
covers the entire space of candidate models, we can bound the overall
regret, as well as bound the number of parameter values we have to
track, and the inherent tradeoff between the two. We propose to choose
a finite collection of parameters from a closed, convex set
$\mathcal{A}$ and perform DFS (Alg.~\ref{alg:dfs}) on this
collection. We specifically consider the case where the true dynamical
model $\alpha^* \in \mathcal{A}$ is unchanging in time and use DFS
with $m=0$. (Fixed share with $m=0$ amounts to the Exponentially
Weighted Averaging Forecaster \cite{Vov90, LitWar94,CesLug06}.) In the
below, for any $\alpha \in \mathcal{A}$, let
$$V_\Phi(\btheta_T,\alpha) \deq \sum_{t=1}^{T-1} \|\theta_{t+1}-\Phi_t(\theta_t,\alpha)\|.$$

\begin{theorem}[Covering sets of dynamics parameter
  space]\label{thm:covering} Let $\varepsilon_N >0$ and
  $\mathcal{A}_N$ denote a covering set for $\mathcal{A}$ with
  cardinality $N$, such that for every $\alpha \in \mathcal{A}$, there
  is some $\alpha' \in \mathcal{A}_N$ such that $\|\alpha - \alpha'\|
  \le \varepsilon_N$. Define candidate dynamical models as
  $\Phi_t(\cdot,\alpha)$ for $\alpha \in \mathcal{A}_N$ and assume
  they are all contractive with respect to the Bregman Divergence used
  in Alg. \ref{alg:dmd}. If $\|\Phi_t(\theta,\alpha) -
  \Phi_t(\theta,\beta)\|\leq L \|\alpha-\beta\|$ for some $L>0$ for
  all $\alpha,\beta \in \mathcal{A}$, then for some constant $C>0$,
  the Alg \ref{alg:dfs} with $\eta_t=\frac{1}{\sqrt{t}}, \eta_r =
  \sqrt{\frac{2 \log(N)}{T}},\lambda=0$ yields a tracking regret
  bounded by
$$C\left(\sqrt{T}\left[\sqrt{\log(N)} +  \min_{\alpha \in \mathcal{A}}V_\Phi(\btheta_T,\alpha)  + T \varepsilon_N \right]\right).$$  
\end{theorem}

Intuitively, we know that if we set $\varepsilon_N$ to be very small
we will have good performance because any possible parameter value
$\alpha \in \mathcal{A}$ would have to be close to a candidate dynamic; however, we
would need to choose many candidates.  Conversely, if we run DFS on
only a few candidate models, it will be computationally much more
efficient but our total regret will grow due to parameter mismatch.

\begin{corollary} Assume $\mathcal{A} \subseteq
  [A_{\min},A_{\max}]^n$, and let $\gamma > 0$ be given.  Let
  $k=\lceil (A_{\max}-A_{\min}) n T^\gamma/2\rceil$ and $\partial =
  (A_{\max}-A_{\min})/(2k)$; let $\mathcal{A}_N =
  \{A_{\min}+\partial,A_{\min}+3\partial,\ldots,A_{\min}(2 k -
  1) \partial\}^n$ correspond to an $n$-dimensional grid with $k^n$
  grid points over $\mathcal{A}$. Then
$$ \max_{\alpha \in \mathcal{A}} \min_{\alpha' \in \mathcal{A}_N} \|\alpha - \alpha'\|_1 \leq T^{-\gamma}.
$$
Additionally, the total number of grid points is upper bounded by
$$N\leq \left(\frac{(A_{\max}-A_{\min} )n T^\gamma}{2}+1\right)^n = O(T^{\gamma n})$$
Under the assumptions of Theorem \ref{thm:covering}, with this set
$\mathcal{A}_N$ and using the fact that
  norms are equivalent on finite-dimensional vectors (\ie there's a
  finite $Z>0$ such that $\|\alpha-\beta\|_1 \leq Z \|\alpha-\beta\|$
  for any $\alpha,\beta \in \mathcal{A}$ for any norm), we get the
following bound on regret for some constant $C>0$.
$$R_T(\btheta_T) \leq C\left( \sqrt{T} \left[ \sqrt{\gamma n \log(T)} + \min_{\alpha \in A}V_\Phi(\btheta_T,\alpha) \right] + T^{1-\gamma n}\right)$$
\end{corollary}
Here we have an explicit tradeoff between regret and computationally
accuracy controlled by $\gamma$, since the computational
complexity is linear in $N = O(T^{\gamma n})$.

We can further control the tradeoff between computation complexity and
performance by allowing $\varepsilon_N$ to vary in time.  This could
be done by using the doubling trick, setting temporary time horizons,
and then refining the grid once the temporary time horizon is
reached using a slightly different
experts algorithm which could account for the
changing number of experts as in \cite{Shal11}.

\subsection{Additive dynamics in exponential families}
\label{sec:additive}
The approach described above for generating a covering set of
dynamical models may be effective when the dimension of parameters is
small; however, in higher dimensions, this approach can require
significant computational resources. In this section, we consider an
alternative approach that only requires the computation of predictions
for a single dynamical model. We will see that in some settings, the
prediction produced by Dynamic Mirror Descent (DMD) and a certain set
of parameters for the dynamic model can quickly be converted to the
prediction for a different set of parameters. While the method
described in this section is efficient and admits strong regrets
bounds, it is applicable only for loss functions derived from
exponential families.

The basics of exponential families are described in
\cite{AmaNag00,WaiJor08}, and Mirror Descent in this setting is
explored in \cite{AzoWar01,raginsky_OCP}. We assume some $\phi : \cX
\to \reals^d$ which is a measurable function of the data, and let
$\phi_k$, $k=1,2,\ldots,d$, denote its components:
$$ \phi(x) = \left( \phi_1(x),...,\phi_d(x) \right)^T. $$
We use the specific loss function
\begin{subequations}
\label{eq:expfam}
\begin{equation}
\ell_t(\theta) =  -\log p_\theta(x_t)
\end{equation}
 where
\begin{equation}
p_\theta(x) \deq \exp\{ \ave{\theta,\phi(x)} - Z(\theta)\}
\end{equation}
\end{subequations}
 for a sufficient statistic $\phi$ and $Z(\theta) \deq \log \int \exp\{\ave{\theta,\phi(x)} \} dx$, known as the {\em log-partition function}, ensures that $p_\theta(x)$ integrates to a constant independent of $\theta$. Furthermore, as in \cite{AzoWar01,raginsky_OCP}, we use the Bregman divergence corresponding to the Kullback-Leibler divergence between two members of the exponential family:
\begin{align*}
D(\theta_1\|\theta_2) = Z(\theta_1) - Z(\theta_2) - \ave{\grad Z(\theta_2), \theta_1-\theta_2}.\end{align*}
In our analysis we will be using the {\em Legendre--Fenchel dual} of $Z$
\cite{HirLem01,BoyVan04}:
$$
Z^*(\mu) \deq \sup_{\theta \in \Theta} \left\{
  \ave{\mu, \theta} - Z(\theta) \right\}.
$$
Let $\Theta^*$
denote the image of $\Int \Theta$ under the gradient mapping $\nabla
Z$ i.e. $\Theta^* = \nabla Z(\Int \Theta)$. 
An important fact is that the gradient mappings $\nabla Z$ and
$\nabla Z^*$ are inverses of one another
\cite{BecTeb03,CesLug06,NJLS09}: \begin{align*}
  \left.  \begin{array}{l}
      \nabla Z^* (\nabla Z(\theta)) = \theta \\
      \nabla Z ( \nabla Z^* (\mu)) = \mu \end{array} \right\} \qquad
  \forall \theta \in \Int \Theta, \mu \in \Int \Theta^* \end{align*} 
  Following
\cite{CesLug06}, we may refer to the points in $\Int \Theta$ as the {\em
  primal points} and to their images under $\nabla Z$ as the {\em dual
  points}. For simplicity of notation, in the sequel we will write $\mu = \grad Z(\theta)$, $\theta = \grad Z^*(\mu)$, $\wh{\mu}_t = \grad Z(\wh{\theta}_t)$, etc. 

Additionally, we will use a dynamical model that takes on a specific form:
\begin{align}
\Phi_t(\theta,\alpha) = \grad Z^*( A_t\grad Z(\theta) + B_t\alpha + c_t) \label{eq:add}
\end{align}
for $\theta \in \Int \Theta, c_t \in \reals^d$, $\alpha \in \reals^n$,
$A_t \in \reals^{d \times d}$, and $B_t \in \reals^{d \times
  n}$. $A_t, B_t$ and $c_t$ are considered known.  
  This dynamical model encompasses some important scenarios. For instance if the log-partition function is the regular $\ell_2$ function, this model includes all dynamics in the form of $\theta_{t+1}=A_t \theta_t + B_t \alpha + c_t$, which is akin to an autoregressive moving average model as in Section \ref{exp:add}. These types of models could also be used to push estimates towards a known temporal structure, e.g. a diurnal pattern with unknown amplitude as in Section \ref{exp:Enron}. Additionally, $B_t$ could encode additional side information. Using these
dynamics, we let $\wh{\theta}_{\alpha,t}$ denote the output of DMD
(Alg.~\ref{alg:dmd})
 at time $t$ and $\wh{\mu}_{\alpha,t}$ be its dual.  Under
all these conditions, we have the following Lemma.

\begin{lemma}\label{lem:induction}
  For any $\alpha,\beta \in \mathcal{A}$, let $\wh{\mu}_{\alpha,1} =
  \wh{\mu}_{\beta,1}$ be the duals of the initial prediction for DMD
  and $K_1=\zeros \in \reals^{d \times n}$. Additionally assume that
  the minimizer of equation \ref{eq:dmd1} is a point in $\Int \Theta$
  for any parameter $\alpha \in \mathcal{A}$.  Then the DMD prediction
  under a dynamical model parameterized by $\alpha$ can be calculated
  directly from the DMD prediction under a dynamical model
  parameterized by $\beta$ for $t>0$ as
\begin{align*}
\wh{\mu}_{\alpha,t} = &\wh{\mu}_{\beta,t} + K_t(\alpha-\beta)\\
{\text{where  }}  K_t =& (1-\eta_{t-1})A_{t-1}K_{t-1} + B_{t-1}.
\end{align*}
\end{lemma}

From Lemma~\ref{lem:induction}, we see that the prediction for
dynamical model $\alpha$ can be computed simply from the prediction
using parameters $\beta$ and the value $K_t$. This is a significant
computational gain compared to DFS, where we had to keep track of
predictions for each candidate dynamical model individually and
therefore needed to bound the number of experts for tractability.

Algorithm~\ref{alg:additive} leverages Lemma~\ref{lem:induction} to
simultaneously track both $\htheta_t$ and the best dynamical model
parameter $\alpha$. In this algorithm, $\wt{\ell}_t$ is the function
defined as
$$\tilde{\ell}_t(\mu) \deq \ell_t(\grad Z^*(\mu)) \equiv
\ell_t(\theta).$$ The basic idea is the following: we use Stochastic Gradient Descent to compute an estimate of the best dynamical model parameter,
compute the DMD prediction associated with that parameter, and then
use DMD to update that prediction for the next round.

\begin{algorithm}[t]
\caption{Dynamic mirror descent (DMD) with parametric additive dynamics}
\label{alg:additive}
\begin{algorithmic} 
\STATE Given decreasing sequence of step sizes $\rho_t,\eta_t > 0$
\STATE Initialize $\wh{\alpha}_1 = \zeros$, $K_1 = \zeros$, $\wh{\theta}_{1} \in \Theta$, $\wh{\mu}_1 = \grad Z(\wh{\theta}_1)$
\FOR {$t=1,...,T$}
\STATE Observe $x_t$
\STATE Incur loss $\ell_t(\wh{\theta}_{t}) = -\ave{\wh{\theta}_{t},\phi(x_t) }+ Z(\wh{\theta}_{t})$
\STATE Set $g_t(\alpha) = \wt{\ell}_t(\wh{\mu}_{\alpha,t}) \equiv \ell_t(\wh{\theta}_{\alpha,t})$
\STATE Set $\wh{\alpha}_{t+1} = {\text{proj}}_{\mathcal{A}}\left(\wh{\alpha}_{t} - \rho_t \grad g_t(\wh{\alpha}_t)\right)$
\STATE Set $\mu'_{t+1} = \wh{\mu}_{t} + K_t (\wh{\alpha}_{t+1}-\wh{\alpha}_t)$
\STATE Set $\wt{\mu}_{t+1} = (1-\eta_t) \mu'_{t+1} + \eta_t \phi(x_t)$
\STATE Set $\wt{\theta}_{t+1} = \grad Z^*(\wt{\mu}_{t+1})$
\STATE Set $\wh{\theta}_{t+1}=\Phi_t(\wt{\theta}_{t+1},\wh{\alpha}_{t+1})$
\STATE Set $K_{t+1} = (1-\eta_t)A_tK_{t} + B_t$ 
\ENDFOR
\end{algorithmic}
\end{algorithm}

\begin{theorem}\label{thm:additive} 
Assume that the observation space $\cX$ is bounded. Let $\Theta \subset \reals^d$ be a bounded, convex set satisfying the following properties for a given constant $H>0$:
\begin{itemize}
\item For all $\theta \in \Theta$,
$$Z(\theta) \deq \int_\cX \exp \big\{{\ave{\theta,\phi(x)}}\big\} d\nu(x) <
+\infty.  $$ 
\item For all $\theta \in \Theta$, $\nabla^2 Z(\theta) \succeq 2H I_{d \times d}.$
\item Let $f_t$ denote the objective function in \eqref{eq:dmd1}. For every $x \in \cX$ and $t \in \{1,2,3,\ldots\}$,  the solution to
 $\displaystyle\argmin_{\theta \in \Theta} f_t(\theta)$ occurs where $\grad f_t = \zeros$.
\end{itemize}
If the assumptions of Lemma \ref{lem:induction} hold, and $\Phi_t(\theta,\alpha)$ is contractive for all $\alpha \in \mathcal{A}$ with respect to the Bregman Divergence induced by $Z(\theta)$, the loss function is of the form \eqref{eq:expfam} and $\wt{\ell}_t(\mu) \deq \ell_t(\grad Z^*(\mu))$ is convex in $\mu$, and $\eta_t,\rho_t \propto 1/\sqrt{t}$ or $1/\sqrt{T}$,
then
the regret for any comparator sequence $\boldsymbol{\theta}_T \in \Theta^T$ associated with Algorithm~\ref{alg:additive} for dynamical models of the form \eqref{eq:add} is
$$
R_T(\btheta_T) \le C \sqrt{T}\left(1 + \min_{\alpha \in \mathcal{A}}
  V_\Phi(\btheta_T,\alpha)\right) 
$$
 for some constant $C>0$.
 \end{theorem}

The condition that $\tilde{\ell}_t(\mu)$ is convex is a sufficient condition to ensure that $g_t(\alpha)$ is convex, which we need in order to search for the optimal value of $\alpha$. While this may not hold true for all exponential family distributions, it holds for many common choices such as Gaussian, Poisson, Binomial, Exponential and many others. Theorem~\ref{thm:additive} shows that Algorithm~\ref{alg:additive}
allows us to simultaneously track predictions and dynamics, and we
perform nearly as well as if we knew the best dynamical models for the
entire sequence in hindsight. While this approach is only applicable
for specific forms of the loss functions and dynamical models, those
forms arise in a wide variety of practical problems. Additionally, this methods allows for a much larger class of comparators which would yield sublinear regret. In fact, we now have an algorithm with sublinear regret for any class of comparators which has a low variation term $\sum_{t=1}^T \|\theta_{t+1} - \Phi_t(\theta_t,\alpha)\|$ for any value $\alpha \in \mathcal{A}$.

\section{Experiments and results}\label{sec:results}

As mentioned in the introduction, many online learning problems can
benefit from the incorporation of dynamical models.  In this section, we
describe how the ideas described and analyzed in this paper might be
applied to many different settings.

\subsection{DMD experiment: dynamic textures with missing
  data} \label{exp:solar}

As mentioned in the introduction, sensors such as the Solar Data
Observatory are generating data at unprecedented
rates. Heliophysicists have physical models of solar dynamics, and
often wish to identify portions of the incoming data which are
inconsistent with their models. This ``data thinning'' process is an
essential element of many big data analysis problems. We simulate an
analogus situation in this section. 

In particular, we consider a datastream corresponding to a dynamic
texture \cite{Szummer96}\cite{dorettoCWS03} , where spatio-temporal dynamics within motion
imagery are modeled using an autoregressive process. In this
experiment, we consider a setting where ``normal'' autoregressive
parameters are known, and we use these within DMD to track a scene
from noisy image sequences with missing elements. (Missing elements
arise in our motivating solar astronomy application, for instance,
when cosmic rays interfere with the imaging detector array.) As
suggested by our theory, the tracking will fail and generate very
large losses when the posited dynamical model is inaccurate. 

More specifically, the idea of dynamic textures is that a low
dimensional, auto-regressive model can be used to simulate a video
which replicates a moving texture such as flowing water, swirling
fog, solar plasma flows, or rising smoke.  This process is modeled in
the following way:
\begin{align*}
\theta_t&=A\theta_{t-1} + B u_t\\
x_t &= C_0 + C \theta_t + D v_t\\
v_t,u_t &\sim \mathcal{N}(0,I).
\end{align*}
In the above, $\theta_t$ denotes the true underlying parameters of the
system, and $x_t$ the observations.  The matrix $A$ is the
autoregressive parameters of the system, which will be unique for the
type of texture desired, $C_0$ the average background intensity, $C$
is the sensing matrix which is usually a tall matrix, and $B$ and $D$
encode the strength of the driving and observation noises
respectively.  Using the toolbox developed in \cite{DTBox} and
samples of a 220 by 320 pixel ocean scene \cite{dorettoCWS03}, we learned
two sets of parameters $A, A' \in \reals^{50\times 50}$, one
representing the water flowing when the data is played forward, and
the other when played backwards, as well as corresponding parameters
$C_0\in \reals^{70400},  C,C' \in \reals^{70400\times 50}, B,B' \in \reals ^{50}$ and $D,D' \in \reals^{70400}$.  Parameters $\theta_t \ \in [-500,500]^{50}$ and data $x_t \in [-500, 500]^{70400}$ were then generated using these parameters, with the parameters $A',B',C',D'$ and $C_0$ on $t=100,...,120$ and $t=300,...,320$ and the parameters $A,B,C,D,C_0$ on the rest of $t=1,...,550$ according to the above equations.  Finally, every observation is corrupted by 50\% missing values, chosen uniformly at random at every time point.  

\begin{figure}[t]
\centering
\includegraphics[height=1.85 in]{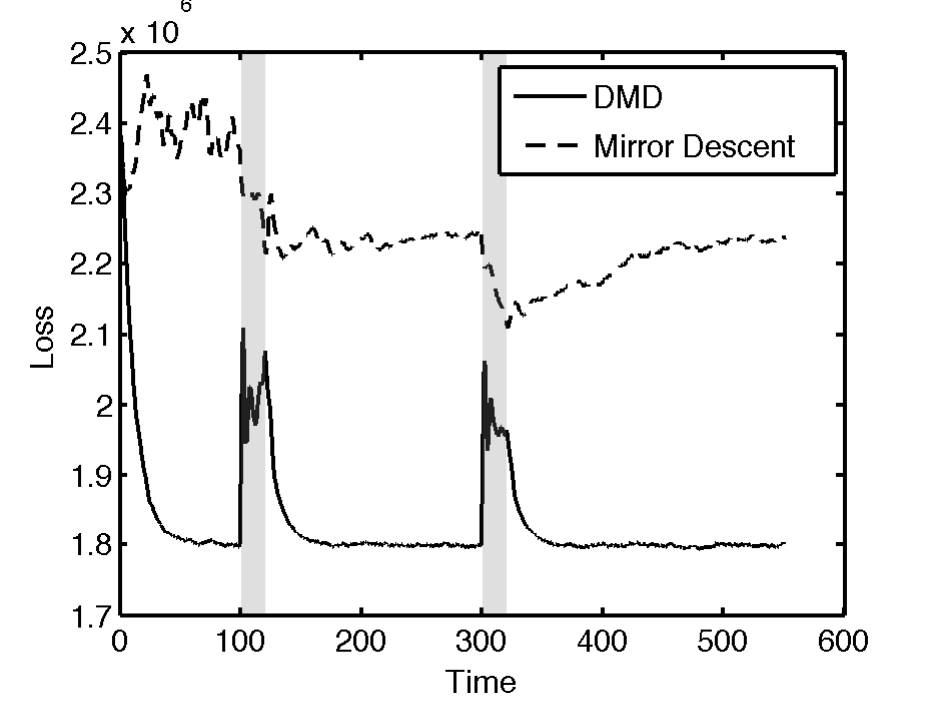}
\caption{Loss curves for the proposed DMD method and classical MD
  vs. time, averaged over 100 trials for the experiment in
  Section~\ref{exp:solar}. The gray areas indicate the intervals where
  the flow being imaged changed direction and hence the posited
  dynamical model was not reflected by the underlying data; note the
  sharp increases in the losses associated with DMD over those
  intervals, particularly in contrast with the losses associated with
  MD. Standard online learning methods like MD do not facilitate the
  detection of time periods with anomalous dynamics. 
}
\label{fig:dynamic_text}
\end{figure}

The parameters $A$, $C_0$, and $C$ were then used to define our
(imperfect) dynamical model for DMD, $\Phi_t(\theta)=A \theta$, and a
loss function $\ell_t(\theta) = \|P_t(C \theta - C_0 - x_t)\|_2^2$,
where $P_t$ is a linear operator accounting for the missing data.
Note that $B$ and $D$ are not reflected in these choices despite
playing a role in generating the data; our theoretical results hold
regardless. We use $\psi(\cdot) =
\frac{1}{2}\|\cdot\|^2_2$ so the Bregman Divergence
$D(x\|y)=\frac{1}{2}\|x-y\|^2_2$ is the usual squared Euclidean
distance, and we perform no regularization
($r(\theta)=0$).  We set $\eta_t = \frac{1}{2\sqrt{t}}$, and ran 100 different trials comparing the DMD
method to regular Mirror Descent (MD) to see the advantage of
accounting for underlying dynamics. The results are shown in Figure
\ref{fig:dynamic_text}.

There are a few important observations about this procedure.  The first
is that by incorporating the dynamic model, we produce an estimate
which visually looks like the dynamic texture of interest, instead of
the Mirror Descent prediction, which looks like a single snapshot of
the water.
 Second, we can recover a good representation
of the scene with a large amount of missing data, due to the
autoregressive parameters being of a much lower dimension than the
data itself.  Finally, because we are using the dynamics of forward
moving water, when the true data starts moving backward, {\em a change
  that is imperceptible visually}, the loss spikes, alerting us of the
abnormal behavior.
\subsection{DMD experiment: solar flare detection with missing data} \label{exp:solar}
We use DMD in order to detect solar flares in image sequences from the Solar Data Observatory,
in the presence of missing data and camera jitter.  Solar flares represent temporally short, and spatially localized bursts of activity 
of the solar scene, but can be hard to detect when there is larger motion, and only partial observations.
By explicitly accounting for camera jitter, we can do a better job of predicting what the scene should look like, and therefore 
quickly observe when and where in the scene solar flares occur.

In order to run DMD we choose an $\ell_2$ loss function on the values of observed pixels
$$\ell_t(\htheta_t)=\frac{1}{2}\|\Omega_t(\htheta_t) - \Omega_t(x_t)\|_2^2$$
where $\Omega_t$ is a function that preserves the value of the true scene at the observed locations and sets the value to 0 otherwise. We use Euclidean distance as our Bregman Divergence. The dynamical model, $\Phi_t$ accounts for the camera jitter at time $t$. For our experiment we chose $\Phi_t$ to be a random pixel shift one pixel up, down, left or right. This scheme is run where each pixel is missing uniformly at random with probability 0.5. These results shown in Figure~\ref{fig:solar}.
\newlength{\figwidth}
\setlength{\figwidth}{0.4\linewidth}
\begin{figure}[t!] 
\begin{center}
\subfloat[Observations with 50\% missing data]{\includegraphics[width=\figwidth]{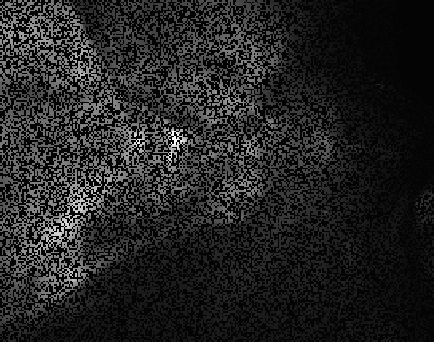}}~
\subfloat[Mirror Descent residual at $t=214$ with 50\% missing data]{\includegraphics[width=\figwidth]{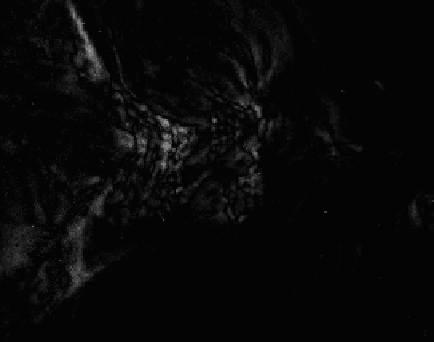}}\\
\subfloat[DMD residual at $t=214$ with 50\% missing data]{\includegraphics[width=\figwidth]{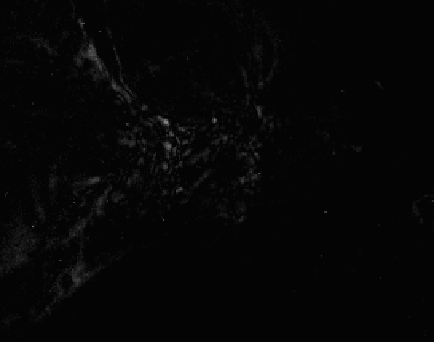}}~
\subfloat[Mirror Descent and DMD losses with 50\% missing data]{\includegraphics[width=\figwidth]{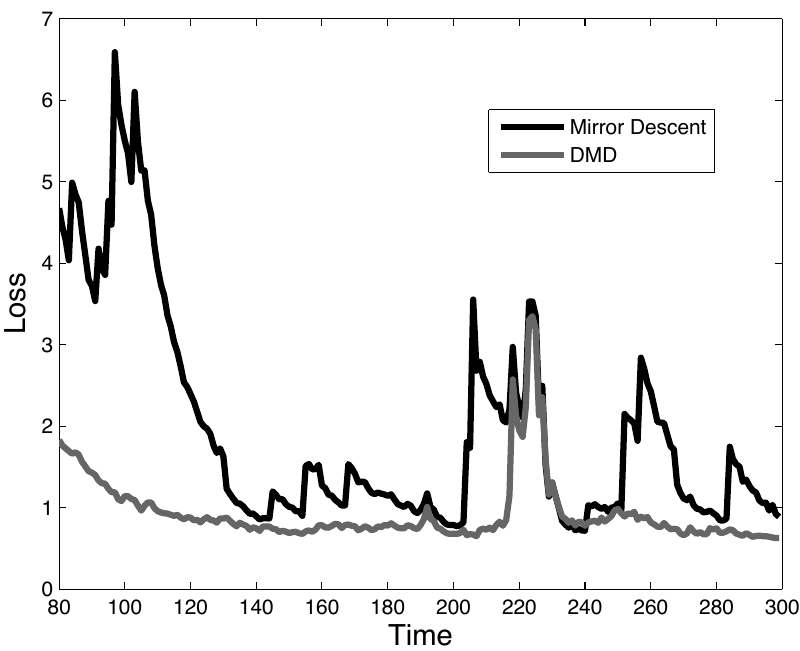}}~
\caption{Results of DMD experiment in Section~\ref{exp:solar}. Images
  display data and residuals with respect to ground truth with missing data. The residuals for the proposed DMD method
  are largest in the locations where a true solar flare occurs, while
  the MD residuals are large in locations exhibiting normal solar
  activity. Furthermore, DMD produces consistently lower, more stable
  losses. Consequently, real anomalous behavior is more easily visible
  at $t=220$ via DMD, while MD produces large spikes in the losses in
  several inactive periods.}
\label{fig:solar}
\end{center}
\end{figure}

We can see in the loss plots that accounting for camera jitter with DMD explicitly we are less likely to get erroneous spikes in the loss function.  This is important, because after the initial learning time, we can threshold the loss to detect anomalies as in \cite{raginsky_OCP}, which would correspond to solar flares in this setting.  However, if we don't explicitly account for the camera jitter, the Mirror Descent loss plot has spikes that are of the same magnitude as the spikes corresponding to a solar flare, and therefore thresholding would lead to either many false positives or missed detections.  

\subsection{DFS experiment: compressive video
  reconstruction}\label{exp:CS}

There is increasing interest in using ``big data'' analysis techniques
in applications like high-throughput microscopy, where scientists wish
to image large collections of specimens. This work is facilitated by
the development of novel microscopes, such as the recent fluorescence
microscope based on structured illumination and compressed sensing principles
\cite{studer2012compressive}. However, measurements in such systems
are acquired sequentially, posing significant challenges when imaging
live specimens.

Knowledge of underlying motion in compressed sensing image sequences
can allow for faster, more accurate reconstruction
\cite{cca_video,WakinVideo,cs-muvi}. By accounting for the underlying
motion in the image sequence, we can have an accurate prediction of
the scene before receiving compressed measurements, and when the
measurements are noisy and the number of observations is far less than
the number of pixels of the scene, these predictions allow both fast
and accurate reconstructions.  If the dynamics are not accounted for,
and previous observations are used as prior knowledge, the
reconstruction could end up creating artifacts such as motion blur or
overfitting to noise.  There has been significant recent interest in
using models of temporal structure to improve time series estimation
from compressed sensing observations \cite{dynamicCS,modifiedCS}; the
associated algorithms, however, are typically batch methods poorly
suited to large quantities of streaming data. In this section we
demonstrate that DMD helps bridge this gap.

In this section, we simulate fluorescence microscopy data generated by
the system in \cite{studer2012compressive} while imaging a paramecium
moving in a 2-dimensional plane; the $t^{\rm th}$ frame is denoted
$\theta_t$ (a $120 \times 120$ image stored as a length-$14400$
vector) which takes values between 0 and 1.  The corresponding
observation is $x_t = A_t\theta_t + n_t,$ where $A_t$ is a $50
\times 14400$ matrix with each element drawn iid from $\mathcal{N}(0,1)$ 
and $n_t$ corresponds to measurement noise with
$n_t \sim \mathcal{N}(0,\sigma^2)$ with $\sigma^2=0.1$. This model
coincides with several compressed sensing architectures
\cite{riceCamera,studer2012compressive}.
\begin{figure}[t]
\centering
\includegraphics[height=1.6
  in]{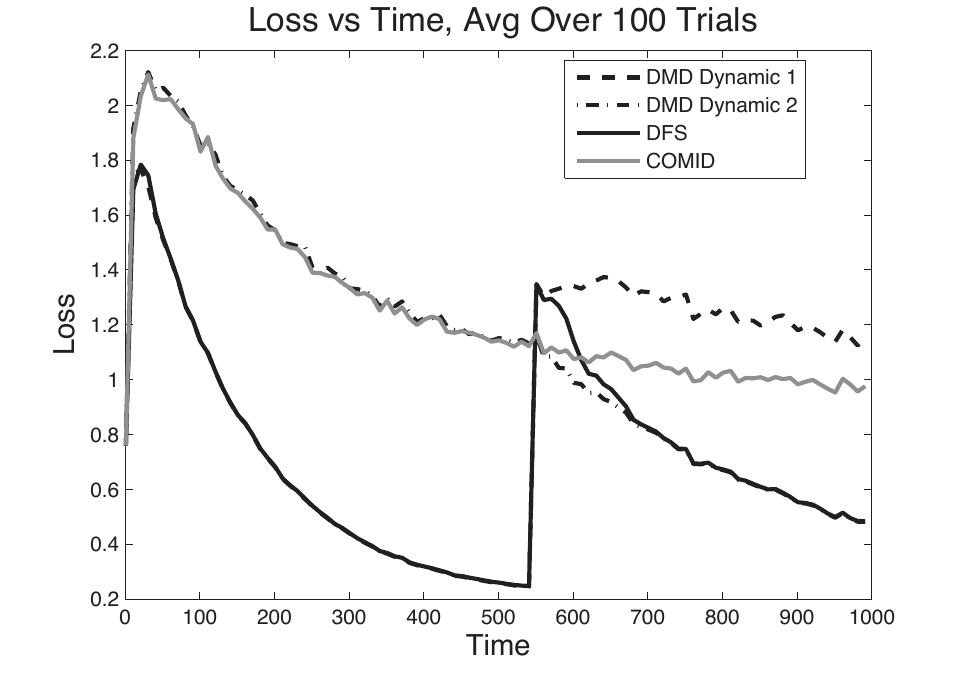}
\caption{Losses averaged over 100 trials using DFS and comparing individual models
  for directional motion for the experiment in
  Section~\ref{exp:CS}. $N=9$ candidate dynamical models were
  considered within the DFS algorithm; plots for the two more accurate
  models are shown for clarity. 
  Before $t=550$ the upward motion dynamic
  model incurs small loss, where as after $t=550$ the motion to the
  right does well, and DFS successfully tracks this change. }
\label{fig:cs}
\end{figure}
\begin{figure}[t]
\centering
\includegraphics[height=2.25in]{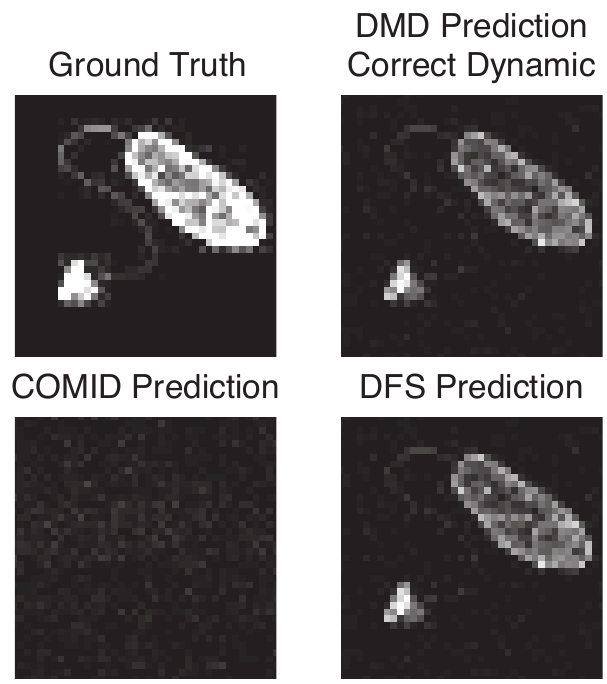}
\caption{Zoomed in instantaneous predictions at $t=1000$ for the
  experiment in Section~\ref{exp:CS}.  Top Left: True $\theta_t$.  Top
  Right: proposed DMD estimate $\htheta_{{\rm{Right}},t}$.  Bottom
  Left: $\htheta_{{\rm{COMID}},t}$.  Bottom Right: proposed DFS
  estimate (without knowledge of true dynamics) $\htheta_t$.  The
  prediction made with the prevailing motion is an accurate
  representation of the ground truth, while the prediction with the
  wrong dynamic is an unclear picture.  The DFS algorithm correctly
  picks out the most accurate dynamical model.}
\label{fig:cs_example}
\end{figure}

Our loss function uses $f_t(\theta) = \frac{1}{2 \sigma^2 d}\|x_t -
A_t\theta\|_2^2$ and $r(\theta) = \tau\|\theta\|_1$, where
$\tau > 0$ is a tuning parameter.  We construct a family of $N =9$
dynamical models, where $\Phi_{i,t} (\theta)$ shifts the (unvectorized) frame,
$\theta$, one pixel in a direction corresponding to an angle of $2\pi
i/(N-1)$ as well as a ``dynamic'' corresponding to no motion. (With
the zero motion model, DMD reduces to COMID.) The true video sequence
uses different dynamical models over $t=\{1,\ldots,550\}$ (upward motion) and $t=
\{551,\ldots,1000\}$ (motion to the right).  Finally, we use $\psi(\cdot) =
\frac{1}{2}\|\cdot\|^2_2$ so the Bregman Divergence
$D(x\|y)=\frac{1}{2}\|x-y\|^2_2$ is the usual squared Euclidean
distance. The DMD sub-algorithms use $\eta_t=\frac{1}{\sqrt{t}}, \tau=.002$ and
the DFS forecaster uses $\lambda=\frac{m}{T-1}=\frac{1}{999}$ and
$\eta_r$ is set as in Theorem \ref{thm:dfs}.  The experiment was then
run 100 times.

Figures~\ref{fig:cs} and~\ref{fig:cs_example} show the impact of using
DFS.  We see that DFS switches between dynamical models rapidly and
outperforms all of the individual predictions, including COMID, used
as a baseline, to show the advantages of incorporating knowledge of
the dynamics.

\subsection{DFS experiment: downsampled traffic surveillance reconstruction}\label{exp:Traffic}
The DFS framework can also be used to reconstruct and predict traffic surveillance data by incorporating the approximately known traffic patterns into the estimation procedure.  For a traffic scene with moving objects, one frame will only be informative for the next frame if the motion of the objects are incorporated into the prediction scheme and thus DFS can be used to improve the prediction process for this type of data. We used the Highway I video from the ATON shadow detection database (http://cvrr.ucsd.edu/aton/shadow/index.html) which contains 440 frames of 240 $\times$ 320 pixel images.  The foreground of this video was then extracted using the inexact Augmented Lagrangian Multiplier method of \cite{Lin10}.  This models the ability of a surveillance camera to do on-board pre-processing of data to remove long term background information and only transmit relevant, transient information. From here the image is blurred by a $7\times7$ Gaussian kernel with width defined by $\sigma=1.75$, downsampled by a factor of 4 in both directions (for an overall downsampling factor of 16), and finally Gaussian white noise is added.  The overall process can be described as follows:
$$ x_t=DH\theta_t + n_t$$
where $H$ encodes the blurring operation and $D$ the downsampling, of the vectorized frame $\theta_t$.  The convolution is normalized such that $\|DH\|_2=1$ to ensure signal strength is maintained.  The white noise has standard deviation of 20.  
The Gaussian noise leads intuitively to using an $\ell_2$ data fit function and because we are estimating foreground objects with background removed we use $\ell_1$ regularization to induce sparsity.  Combining these leads to the following loss function:
$$\ell_t(\htheta_t)= \frac{1}{2d} \|DH\htheta_t - x_t\|_2^2 + \frac{\tau}{d} \|\htheta_t\|_1$$
where $d$ is the dimension of $\htheta_t$ and $\tau$ is the parameter which controls the relative amount of sparsity in the solution.

In order to implement DFS, we need a collection of feasible dynamics which can describe the data.  The video shows cars moving generally from the top of the screen to the bottom at a relatively constant speed, so 5 different dynamical models are postulated which are simple row shifts of 10, 14, 18, 22 and 26 pixels respectively.  One slightly more complicated dynamical model is also used which is based on a Block Matching Algorithm.  For this dynamical model, a training set of the first 20 frames of the full data was used. The idea behind the Block Matching Algorithm is that for every $8\times 8$ block in a given frame, a search is done over the entire previous frame for the block which best matches.  For our purposes, we define the values of a block in the image after applications of the dynamics, $\Phi_t(\theta_t)$, to the be the values of the block in $\theta_t$ which most consistently predicted the block of interest in the original training set. This process is defined below, where $\theta$ represents the 240 $\times$ 320 images, and the subscripts $[i:i+7,j:j+7]$ denote the 8 $\times$ 8 block starting at location $i,j$. 
\begin{align*}
\Phi_t(\theta)_{[i:i+7,j:j+7]} &= \theta_{[l(i,j):l(i,j)+7,m(i,j):m(i,j)+7]}\\
l(i,j), m(i,j) &=\\
 \argmin_{l,m} &\sum_{\tau=1}^{19}\|\theta_{\tau+1,[i:i+7,j:j+7]} - \theta_{\tau,[l:l+7,m:m+7]}\|^2
\end{align*}
This map was then smoothed slightly to ensure that objects in one frame remained intact after the application of the dynamics and not distorted.  This procedure produced a dynamical model which mostly showed downward motion, similar to the pixel shifts, but allowed for some horizontal motion as well as allowing slightly different velocities in different regions of the image. Overall this brings the number of candidate dynamical models to $N=6.$

\begin{figure}[t]
\centering
\includegraphics[height=1.75 in]{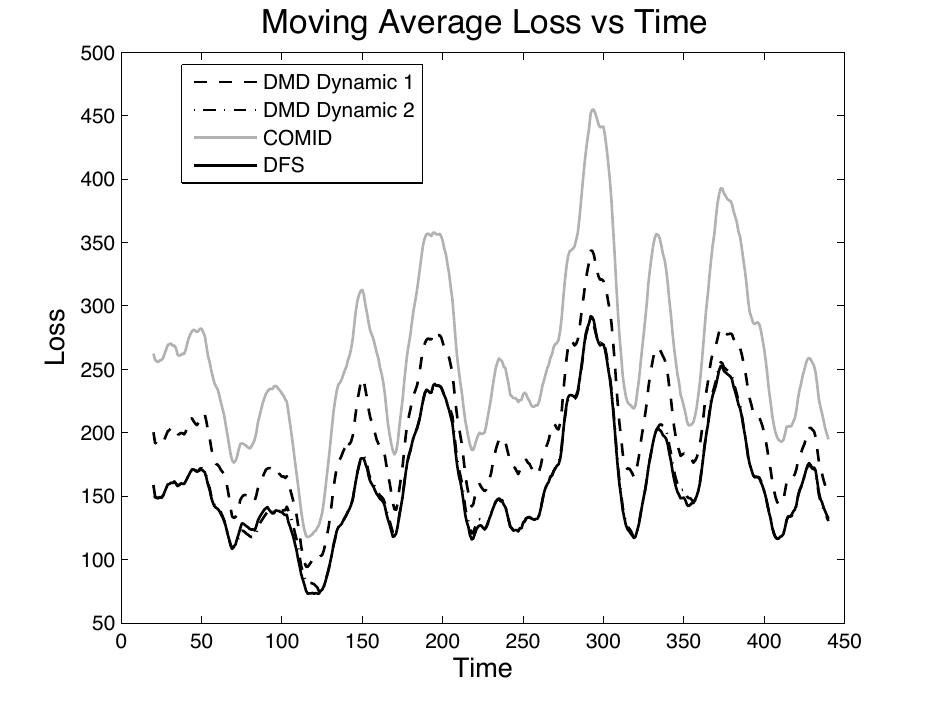}
\caption{Moving average loss with time window=20 of DFS algorithm in the downsampled, noisy traffic surveillance experiment described in Section~\ref{exp:Traffic}.  Only shown are DMD losses for the best pixel shift dynamic and the block matching dynamic for clarity.  Notice that the pixel shift consistently out performs COMID, and the block matching dynamic does even better. DFS quickly finds the best dynamic and follows it closely.}
\label{fig:Traffic_DFS}
\end{figure}

Using this setup, DFS is implemented for the $T=440$ frames with $\tau=10, d=240\times320=76800,\eta_t=\frac{10d}{\sqrt{T}},  \eta_r=\sqrt{\frac{8\left(3\log(N)+2\log(T)+1\right)}{T}},$ and $\lambda=\frac{2}{T-1}$. The results are shown in Figure \ref{fig:Traffic_DFS}. We immediately see that COMID which does not account for dynamics performs very poorly and is easily outperformed even by the very simple pixel shift models (DMD Dynamic 1 in the plots corresponds to pixel shift of 18). Additionally we see that the block matching dynamic (DMD Dynamic 2) performs even better, and that the DFS algorithm quickly hones in on this dynamic model and follows it very closely.  

\begin{figure}[t]
\centering
\subfloat[Ground Truth.]{\includegraphics[height=.85 in]{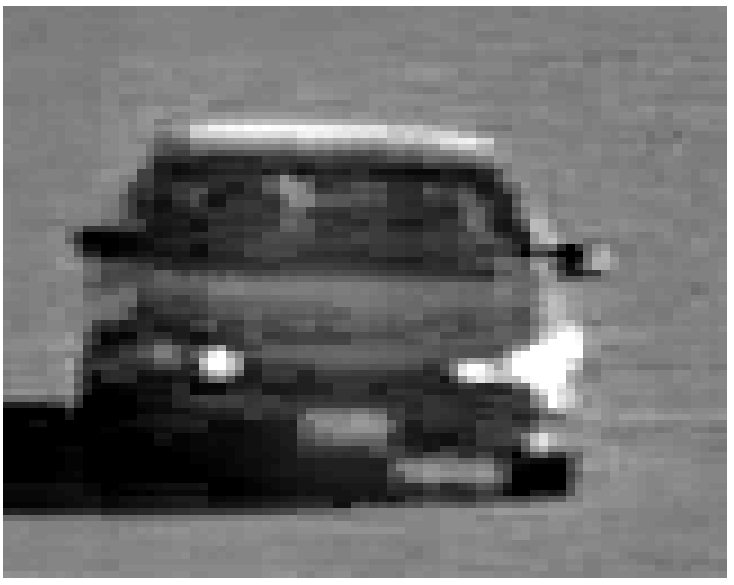}}~
\subfloat[Downsampled Noisy Observation.]{\includegraphics[height=.85
  in]{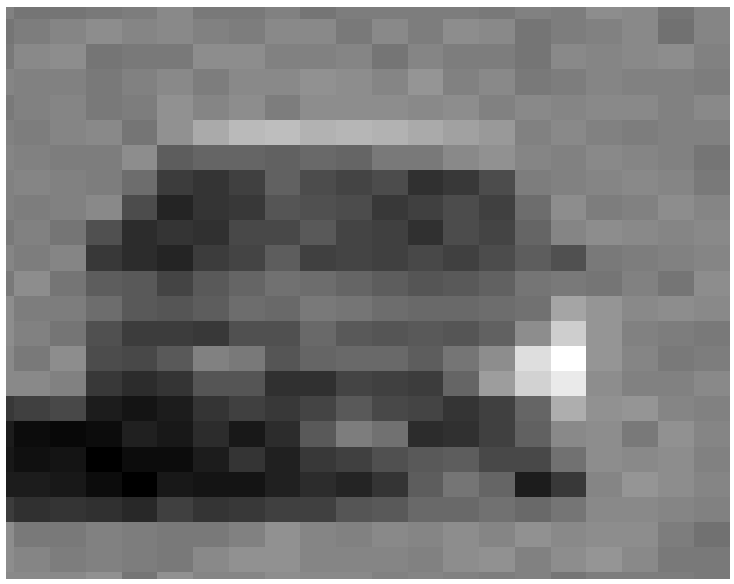}}\\
  \subfloat[COMID Prediction.]{\includegraphics[height=.85
  in]{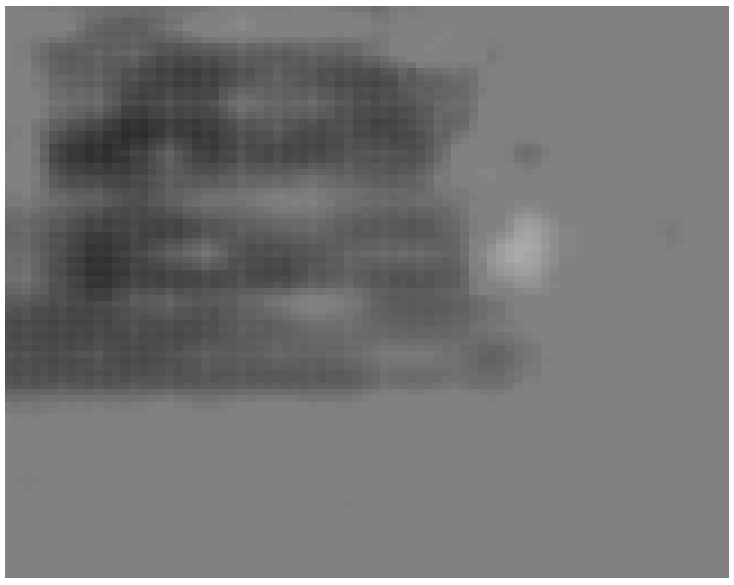}}~
  \subfloat[DFS Reconstruction.]{\includegraphics[height=.85 in]{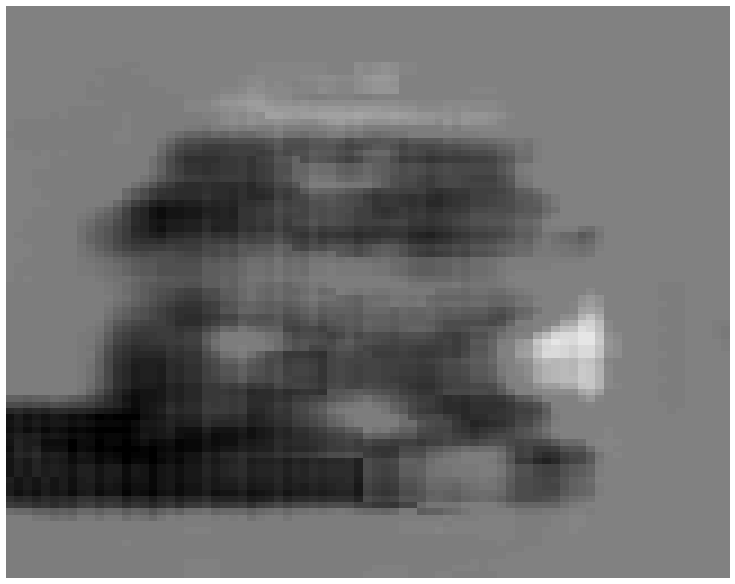}}
\caption{Zoomed in reconstruction of DFS algorithm in the downsampled, noisy traffic surveillance experiment described in Section~\ref{exp:Traffic}. Here we see the ground truth image, observations, COMID prediction and DFS prediction.}
\label{fig:Traffic_Examples_Zoom}
\end{figure}

In addition to the loss plots, we can observe actual reconstructions to see how close DFS comes to approximating the true image. An example of one segment of one prediction is shown in Figure \ref{fig:Traffic_Examples_Zoom}.  We see how the DFS procedure effectively upsamples, deblurs and denoises the data.  By incorporating the dynamics, we can much more effectively find the details in the image that is not possible by using just COMID.  Additionally, we have revealed details unobservable in the raw data.  For instance, notice the lighter shade object inside the car in The lighter shade is visible in the DFS reconstruction, while it is impossible to see in the data. Additionally, details such as shape and edges on the headlights and license plate are much clearer in the DFS reconstruction compared to the data.

\subsection{DMD with parametric additive dynamics}\label{exp:add}

We look at self-exciting point processes on connected
networks \cite{HellerHawkes,ryanAdamsHawkes}. Here we assume
there is an underlying rate for nodes in a network which dictate how
likely each node is to participate in an action. Then, based on which
nodes act, it will increase other nodes likelihood to act in a dynamic
fashion. For example, in a social network a node could correspond to a
person and an action could correspond to crime \cite{BertozziHawkes}.
 In a biological neural network, a node could correspond
to a neuron and an action could correspond to a neural spike
\cite{brown2004multiple}.

We simulate observations of a such a self-exciting point process
in the following way:
\begin{align*}
\mu_{t+1}=\Phi_t(\mu_t,W)=&\tau \mu_t + W x_t + (1-\tau) \bar{\mu}\\
x_t \sim& {\text{Poisson}}(\mu_t)
\end{align*}
For our experiments $\mu_t \in (0,5]^{100}$ represents the average
number of actions each of 100 nodes will make during time interval
$t$, and $W\in [0, 5]^{100 \times 100}$ reflects the unknown underlying
network structure which encodes how much an event by a one node
will increase the likelihood of an event by another node in future
time intervals. Here we assume $\tau$ is a known
parameter between zero and one, $\bar{\mu} \in \reals^{100}$ is a
underlying base event rate.

Our goal is to track the event rates $\mu_t$ and the network model $W$
simultaneously; Algorithm~\ref{alg:additive} is applied with
\begin{align*}
\ell_t(\theta)=&\langle \ones,\exp(\theta) \rangle - \langle
x_t, \theta \rangle, &
\tilde{\ell}_t(\mu)=&\langle \ones,\mu \rangle - \langle x_t,\log \mu
\rangle, \\
Z(\theta)=&\langle \ones, \exp(\theta)\rangle, & \mu =& \grad Z(\theta) =
\exp(\theta). 
\end{align*}

\begin{figure}
  \centering
  \subfloat[ Moving average loss over previous 100 time points for DMD
  with a known $W$ matrix, MD, and DMD exp averaged over 1000
  trials.]{\includegraphics[height=1.75in]{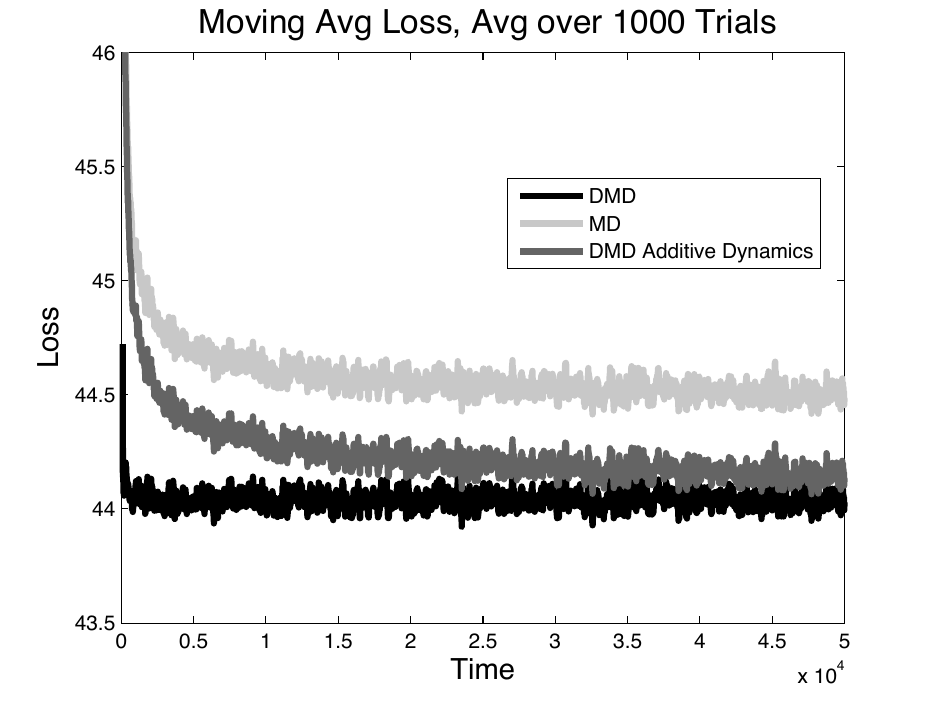}}\\
  \subfloat[The true value and final estimate of $W$ computed using
  Alg~\ref{alg:additive}.]{\includegraphics[height=1.3in]{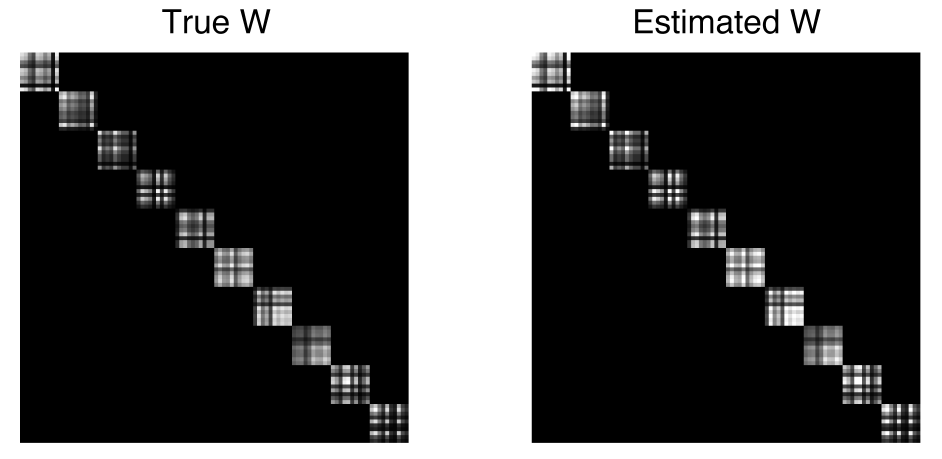}}
  \caption{
    Experimental results tracking a self-exciting point process on a
    network, described in Section~\ref{exp:add}. Notice how the loss
    curve for Alg.~\ref{alg:additive} approaches the DMD curve
    (associated with clairvoyant knowledge of the underlying network
    matrix $W$) as the estimate of $W$ improves, and significantly
    outperforms conventional mirror descent. }
\label{fig:neural_loss}
\end{figure}

We generated data according to this model for $t=1,...,50000$ for 1000
different trials, using $\tau=0.5, \bar{\mu}=0.1$ and $W$ generated
such that it is all zeros except on each distinct $10\times10$ block along the diagonal,
elements are chosen to be $uu^T$ for a vector $u\in[0.1,1.1]^{10}$
with elements chosen uniformly at random.
 The matrix $W$ is then
normalized so that its spectral norm is $0.25$ for stability.  Using
this generated data we ran DMD with known $W$ (Alg.~\ref{alg:dmd}),
MD, and DMD with additive dynamics (Alg.~\ref{alg:additive}) to learn
the dynamic rates.  The step size parameters
were set as $\eta_t= .9/\sqrt{t}$ and $\rho_t=.005/\sqrt{t}$.  
The results are shown for DMD with the matrix $W$
known in advance, MD and Alg.~\ref{alg:additive} in
Figure \ref{fig:neural_loss}.

We again see several important characteristics in these plots.  The
first is that by incorporating knowledge of the dynamics, we incur
significantly less loss than standard Mirror Descent.  Secondly, we
see that even without knowing what the values of the matrix $W$, we
can learn it simultaneously with the rate vectors $\mu_t$ from
streaming data, and the resulting accurate estimate leads to low loss
in the estimates of the rates. 

\subsection{DMD with parametric additive dynamics experiment - Enron email corpus}\label{exp:Enron}
For our final set of experiments, we analyzed the Enron email corpus\cite{Enron:04}. The goal of analyzing this dataset is to be able to read in the emails in a streaming fashion, and be able to detect anomalous moments, and determine if they align with important known events in the company's history.  In order to do this, the approximately 500,000 emails were combined into 1 hour time bins for each of the 150 given employees, from December 16th 2000 until January 1st 2003. For each employee for each hour, we receive an indicator saying whether that employee sent or received an email at that specific hour.  

We model the indicator values as a Bernoulli random variable for each employee for each hour, and want to learn the probability that each employee will send or receive an email at a given time.  Given an estimate $p_t\in[0,1]^{150}$ and the actual observations $x_t \in \{0,1\}^{150}$ at time $t$, we have the following loss function
$$ \tilde{\ell}_t(\hat{p}_t) =- \langle x_t,\log \hat{p}_t \rangle - \langle \ones - x_t, \log \ones - \hat{p}_t \rangle  $$
 which can be rewritten as
 $$ \ell_t(\htheta_t)=-\langle x_t, \htheta_t \rangle + \left\langle \ones,\log\left( \ones + \exp{\htheta_t}\right) \right\rangle$$
 under the transformation $\theta = \log\left(\frac{p}{\ones-p}\right)$.  This formulation gives us the primal ($\theta$) and dual ($p$) variables and loss functions necessary for DMD with parametric additive dynamics algorithm described in Alg. \ref{alg:additive}.
 
 In order to use Alg. \ref{alg:additive} on this data set we need to postulate a sequence of dynamics, and this dataset lends itself to several possibilities.  The first possibility is to search for possible relationships in the network that might effect a person's likelihood to send or receive and email.  For this we consider the following dynamic model:
 $$ \Phi_t(p) = p + \eta_t (W - I) x_t \hspace{.1 in} {\rm{s.t.}} \hspace{.1 in} W_{i,j}\geq 0, \|W\|_{\infty}\leq 1.$$
 This dynamic model might look a little bit strange at first, but when used in Alg. \ref{alg:additive} we are left with the overall update equation of the form
 $$ \hat{p}_{t+1}=(1-\eta_t) \hat{p}_t + \eta_t \widehat{W}_t x_t$$
 where we can search for $\widehat{W}_t$ over the space $\mathcal{W}=\{W {\text{ s.t. }} W_{i,j}\geq 0, \|W\|_\infty \leq 1\}$. This update equation shows that we not only update our estimate of an individual's likelihood of sending or receiving an email based on their previous history, but also the history of people they might be associated with in the network. For instance, if someone is working closely with another employee, they will be more likely to respond quickly to that person's email as opposed to someone who works in another department.
 
 The next two dynamical models attempt to incorporate the weekly cycles in email sending behavior. The first dynamical model incorporates a weighted average over each employee's behavior at corresponding hours in previous weeks to predict their behavior at the current moment, and the second model incorporates a weighted average of the entire company's behavior in previous weeks.  The first model uses the following function:
 $$\Phi_t(p) = p + \eta( (\alpha_0 - 1) x_t + \alpha_1 x_{t+1-K} + \alpha_2 x_{t+1-2K} + ... +\alpha_M x_{t+1-MK})$$
 where $K$ corresponds to the amount of time backwards we wish to look, in this case $24\times 7$ hours, and $M$ is how many of these periods back we wish to look.  This leads to the total update equation
 $$\hat{p}_{t+1}=(1-\eta_t)p_t + \eta_t (\alpha_0 x_t + \alpha_1 x_{t+1-K} + ... + \alpha_M x_{t+1-MK})$$
 where we can use Alg \ref{alg:additive} to search for optimal values of $\alpha$ over the $(M+1)$ dimensional simplex.  A similar dynamical model leads to the final update equation
 \begin{align*}
 \hat{p}_{t+1}=(1-\eta_t) p_t +& \\
 \eta_t\Big(\alpha_0 x_t +& \alpha_1 \frac{\ones\ones^T x_{t+1-K}}{d}+ ... + \alpha_M \frac{\ones\ones^T x_{t+1-MK}} {d}\Big)
\end{align*}
which uses the same mechanism for incorporating weekly patterns, but considers the company-wide information from the previous weeks, instead of the employee specific information. Once again our algorithm allows us to use this update equation while searching for optimal values of $\alpha$ over the $(M+1)$ dimensional simplex. For our experiments, we considered $M$=20.

\begin{figure}
  \centering
  \includegraphics[height=2in]{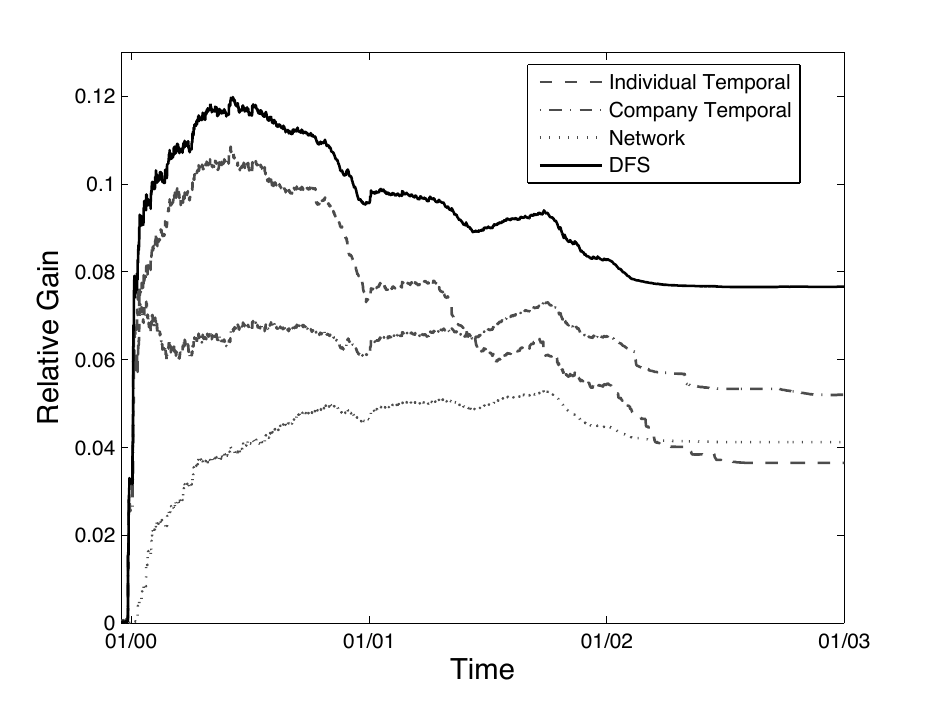}
  \caption{Relative gain of our method's prediction using dynamical models compared to Mirror Descent. Notice that all our proposed dynamics improve performance compared to Mirror Descent in the 4-8\% range at the end of the data set, and at a maximum of around 12\%.}
\label{fig:Enron_results}
\end{figure}

Using this loss function and dynamical models, Alg. \ref{alg:additive} was used to predict each employees probability of sending or receiving an email for every hour in the given time range.  The DMD step size was set as $\eta_t=\frac{10}{\sqrt{T}}$, where $T$ is the total number of hours observed.  Additionally, the step sizes $\rho_t$ were set at $\rho_t=\frac{10^{-9}}{\sqrt{T}}$ for the network dynamics, and $\rho_t=\frac{10^{-1}}{\sqrt{T}}$ for the temporal behavior dynamics. In addition to these three models, we also used Alg. \ref{alg:dfs} to assign weights to each of these methods, and regular Mirror Descent, to create one combined prediction. We compared the performance of each of these methods to Mirror Descent, and calculated the relative gain through time $\tau$ as $\frac{\sum_{t=1}^\tau \ell_t(\htheta_t^{\text{MD}}) - \ell_t(\htheta_t)}{\sum_{t=1}^\tau \ell_t(\htheta_t^{\text{MD}})}$, where $\htheta_t^{\text{MD}}$ corresponds to the prediction made by Mirror Descent. These results are shown in Figure \ref{fig:Enron_results}.

\begin{figure}
  \centering
  \subfloat[Comparison of instantaneous loss versus average of
  preceding week for Mirror Descent.]{\includegraphics[width=.24\textwidth]{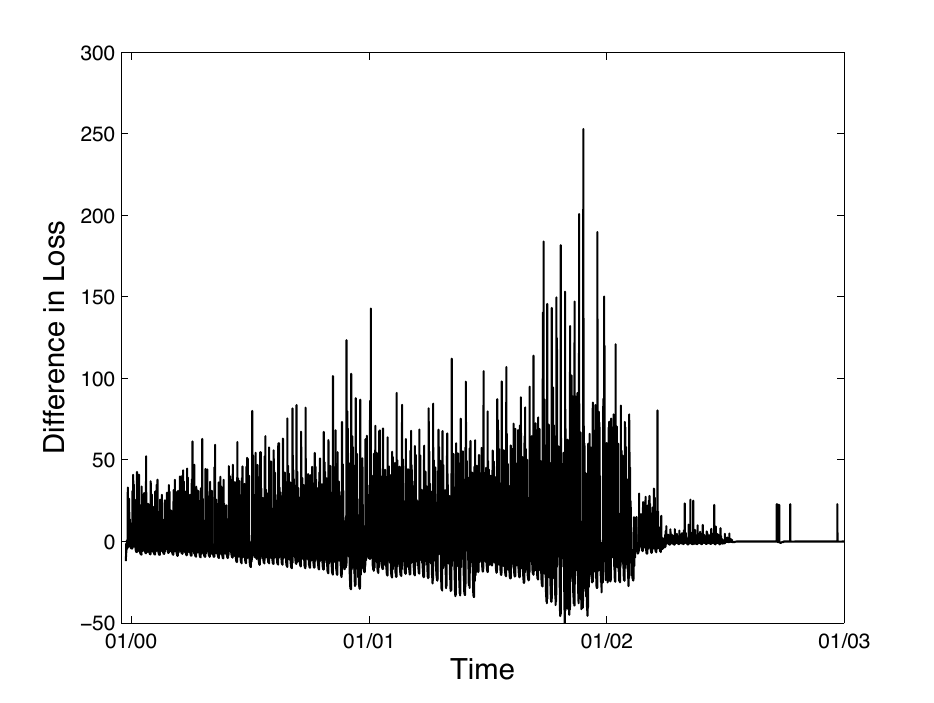}}~
  \subfloat[Comparison of instantaneous loss versus average of preceding week for our DFS method.]{\includegraphics[width=.24\textwidth]{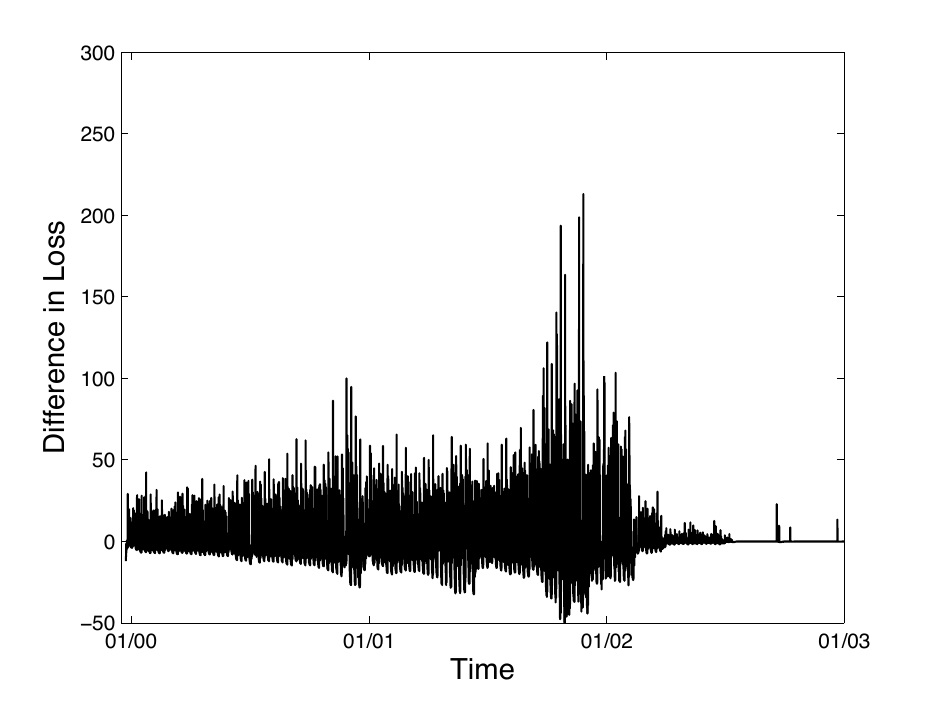}}
  \caption{Filtered loss vs. time plots for the Enron experiment using
    (a) losses stemming from MD and (b) losses stemming from the
    proposed DFS method.  These two plots show that our
    method has larger relative spikes around December 2000 and December 2001
    corresponding to significant moments in the company's history.}
\label{fig:Enron_spikes}
\end{figure}

Immediately we see that all of the proposed dynamic models improve the our predictive power compared to Mirror Descent.  Including individual employees weekly email behavior has the biggest single impact, followed by the company's weekly behavior, and finally the network dynamics. When all three are combined using fixed share, we attain a performance boost of about 8\% in the end, and 12\% at its peak. Additionally, our method can more accurately detect anomalies. One heuristic for determining anomalies would be to compare a method's instantaneous loss to the average loss for the preceding week. High spikes would then correspond to anomalous moments. This value is shown in Figure \ref{fig:Enron_spikes} for Mirror Descent and our DFS method. We see that our method does a better job of finding the spikes in December 2000 and December 2001, corresponding to true anomalies, as the spikes stick out more from the noise around them. In December 2000 energy commodity trading became deregulated in California \cite{pbs_timeline} and in December 2001 Enron filed for Chapter 11 \cite{Enron_bankrupt}.

\section{Conclusions and future directions}\label{sec:conc}
Processing high-velocity streams of high-dimensional data is a central
challenge to big data analysis. Scientists and engineers continue
to develop sensors capable of generating large quantities of data, but
often only a small fraction of that data is carefully examined or
analyzed. Fast algorithms for sifting through such data can help
analysts track dynamic environments and identify important subsets of
the data which are inconsistent with past observations. 

In this paper we have proposed a novel online optimization method,
called Dynamic Mirror Descent (DMD), which incorporates dynamical
models into the prediction process and yields low regret bounds for
broad classes of comparator sequences. The proposed methods are
applicable for a wide variety of observation models, noise
distributions, and dynamical models.  There is no assumption within
our analysis that there is a ``true'' known underlying dynamical
model, or that the best dynamical model is unchanging with time. The
proposed Dynamic Fixed Share (DFS) algorithm adaptively selects the
most promising dynamical model from a family of candidates at each
time step.  Additionally we show methods which learn in parametric
families of dynamical models.  In experiments DMD shows strong
 tracking behavior even when underlying
dynamical models are switching, in such applications as dynamic texture analysis,
 compressive video, and self-exciting point process analysis.
\appendix
\section{Proofs}\label{sec:proofs}
\subsection{Proof of Theorem~\ref{thm:main}}
The proof of Theorem \ref{thm:main} shares some ideas with the
tracking regret bounds of \cite{Zin03}, but uses properties of the
Bregman Divergence to eliminate some terms, while additionally
incorporating dynamics. We employ the following lemma.
\begin{lemma}\label{lem:main}
Let the sequence $\bhtheta_T$ be as in Alg.~\ref{alg:dmd}, and let
$\btheta_T$ be an arbitrary sequence in $\Theta^T$; then
\begin{align*}
\ell_t(\htheta_t)-\ell_t(\theta_t) \leq&
\frac{1}{\eta_t}\left[D(\theta_{t}\|\htheta_t) -
  D(\theta_{t+1}\|\htheta_{t+1}) \right] \\
  &+\frac{\Delta_{\Phi}}{\eta_t}+
\frac{2M}{\eta_t}\|\theta_{t+1}-\Phi_t(\theta_{t})\| +
\frac{\eta_t}{2\sigma}G^2.
\end{align*}
\end{lemma}
\label{pf:lem}
{\bf Proof of Lemma~\ref{lem:main}:}
The optimality condition of \eqref{eq:dmd1} implies
\begin{align}
\ave{\grad f_t(\htheta_t)+&\grad
  r(\ttheta_{t+1}),\ttheta_{t+1}-\theta_{t}} \leq \nonumber\\
\frac{1}{\eta_t}& \ave{\grad \psi(\htheta_t) -
  \grad \psi(\ttheta_{t+1}), \ttheta_{t+1}-\theta_{t}}.
\label{eq:firstord}
\end{align}
The proof has a similar structure to that in \cite{COMD}
\begin{subequations}
{\allowdisplaybreaks
\begin{align}
f_t(&\htheta_t)-f_t(\theta_t)+r(\htheta_{t})-r(\theta_{t}) \\
= &f_t(\htheta_t)-f_t(\theta_t)+r(\htheta_t) - r(\ttheta_{t+1}) + r(\ttheta_{t+1})-r(\theta_{t}) \\
\leq&  \ave{ \grad f_t(\htheta_t), \htheta_t - \theta_t } 
+ \ave{ \grad r(\htheta_t),\htheta_t-\ttheta_{t+1} } \nonumber \\
&+ \ave { \grad r(\ttheta_{t+1}),\ttheta_{t+1} - \theta_t } \label{eq:useConvexity}\\
\le& \ave{\grad f_t(\htheta_t) + \grad r(\ttheta_{t+1}),\ttheta_{t+1}-\theta_t } \nonumber \\
&+ \ave{\grad f(\htheta_t) + \grad r (\htheta_t),\htheta_t - \ttheta_{t+1}}\\
\le& \frac{1}{\eta_t} \ave{\grad \psi(\htheta_t) - \grad \psi (\ttheta_{t+1}), \ttheta_{t+1}-\theta_{t}} \nonumber\\
&+ \ave{\grad f_t (\htheta_t) + \grad r (\htheta_t), \htheta_t -\ttheta_{t+1}} \label{eq:useFirst} \\
 =&\frac{1}{\eta_t} \left( D(\theta_t\|\htheta_t) - D(\theta_t\|\ttheta_{t+1}) - D(\ttheta_{t+1}\|\htheta_t) \right)\nonumber\\
&+ \ave{\grad f_t (\htheta_t) + \grad r (\htheta_t), \htheta_t-\ttheta_{t+1}} \label{eq:useTriBreg2}\\
 =& \frac{1}{\eta_t} \left[D(\theta_{t}\|\htheta_t) -
  D(\theta_{t+1}\|\htheta_{t+1})\right] \nonumber \\
&+ \frac{1}{\eta_t} \underbrace{ \left[  D(\theta_{t+1}\|\htheta_{t+1}) -
  D(\Phi_{t}(\theta_{t})\|\htheta_{t+1})\right] }_{T_1} \nonumber\\
   &+ \frac{1}{\eta_t} \underbrace{\left[
  D(\Phi_{t}(\theta_{t})\|\htheta_{t+1})  -
  D(\theta_{t}\|\ttheta_{t+1}) \right]}_{T_2} \nonumber \\
  &-\underbrace{ \frac{1}{\eta_t} D(\ttheta_{t+1}\|\htheta_t)
 + \ave{ \grad f_t(\htheta_t) + \grad r(\htheta_t),\htheta_t-\ttheta_{t+1}}}_{T_3} \nonumber
  \label{eq:useTriBreg3}
\end{align}}
\end{subequations}
\noindent where \eqref{eq:useConvexity} follows from the convexity of $f_t$ and
$r$, \eqref{eq:useFirst} follows from the optimality condition in \eqref{eq:dmd1}, and
\eqref{eq:useTriBreg2} follows from \eqref{eq:triBreg}.
Each of these terms can be bounded individually, and then recombined to complete the proof.
\begin{subequations}
\begin{align}
T_1  =&  \psi(\theta_{t+1}) -\psi(\Phi_{t}(\theta_{t}))  - \ave{ \grad
\psi(\htheta_{t+1}),\theta_{t+1}-\Phi_{t}(\theta_t)} \\
\leq&  \ave{\grad \psi (\theta_{t+1})-\grad \psi(\htheta_{t+1}),\theta_{t+1}-\Phi_{t}( \theta_t)} \label{eq:PsiConv}\\
\le&\|\grad \psi (\theta_{t+1})-\grad \psi(\htheta_{t+1})\|_* \|\theta_{t+1} - \Phi_{t}(\theta_{t})\|\label{eq:Cauch} \\
\le& 2M\|\theta_{t+1} - \Phi_{t}(\theta_{t})\|\\
T_2 =& D(\Phi_{t}(\theta_t)\|\Phi_{t}(\ttheta_{t+1}))-D(\theta_t\|\ttheta_{t+1})\leq \Delta_{\Phi} \\
T_3  \leq& -\frac{\sigma}{2\eta_t} \| \ttheta_{t+1}-\htheta_t\|^2 +
\|\grad f_t(\htheta_t) + \grad r (\htheta_t) \|_* \|\htheta_t- \ttheta_{t+1} \| \label{eq:StrongBreg}\\
\leq&-\frac{\sigma}{2\eta_t} \| \ttheta_{t+1}-\htheta_t\|^2 
+\frac{\sigma}{2\eta_t} \|\ttheta_{t+1} - \htheta_t\|^2 +
 \frac{\eta_t}{2\sigma} G^2, \label{eq:young}
\end{align}
\end{subequations}
where \eqref{eq:PsiConv} is due to the convexity of $\psi$ and
\eqref{eq:Cauch} is from the Cauchy-Schwarz inequality. Additionally,
\eqref{eq:StrongBreg} is due to \eqref{eq:convBreg} and
\eqref{eq:young} uses Young's Inequality \cite[Prob 9.1]{Steele} which states $ab\leq \frac{a^2}{2\epsilon}+\frac{b^2 \epsilon}{2}$. Combining these
inequalities with \eqref{eq:useTriBreg3} gives the lemma as it is
stated. \hfill $\Box$

The proof of the theorem concludes by
summing the bounds of Lemma~\ref{lem:main} over time.  Denote
$D_t\deq D(\theta_{t}\|\htheta_{t})$ and $V_t \deq \|\theta_{t+1} -
\Phi_t(\theta_t)\|$. 
Remember, we have assumed that $\Delta_\Phi\leq 0$.
{\allowdisplaybreaks
\begin{align*}
R_T(\btheta_T) \le&  \sum_{t=1}^{T}
\left(\frac{D_t}{\eta_{t}}-\frac{D_{t+1}}{\eta_{t+1}}\right) +  \frac{G^2}{2\sigma}\sum_{t=1}^{T} \eta_t \\
&+ D_{\max}\sum_{t=1}^{T}
\left(\frac{1}{\eta_{t+1}}-\frac{1}{\eta_t}\right)  + \sum_{t=1}^{T}
\frac{2M}{\eta_t}V_t
\\
\leq & \frac{D_1}{\eta_1} - \frac{D_{T+1}}{\eta_{T+1}}+D_{\max}\left(\frac{1}{\eta_{T+1}}- \frac{1}{\eta_1}\right) \\
&+\frac{2M}{\eta_T}V_{\Phi}(\btheta_T)+\frac{G^2}{2\sigma}\sum_{t=1}^T\eta_t\\
\le & \frac{D_{\max}}{\eta_{T+1}} +
\frac{2M}{\eta_T}V_{\Phi}(\btheta_T) + 
\frac{G^2}{2\sigma}\sum_{t=1}^{T} \eta_t. \qquad \Box
\end{align*}}
\subsection{Proof of Theorem~\ref{thm:dfs}}
The tracking
regret can be decomposed as:
\begin{align} R_T&(\btheta_T) = \underbrace{\sum_{t=1}^{T} \ell_t(\htheta_t) -
  \min_{\substack{i_1,\ldots,i_T\\ \sum_{t=1}^{T-1}\mathbf{1}[i_t\neq i_{t+1}]\leq m}} \sum_{t=1}^{T}
  \ell_t\left(\htheta_{i_t,t}\right)}_{T_4}\nonumber\\
+& \underbrace{\min_{\substack{i_1,\ldots,i_T\\ \sum_{t=1}^{T-1}\mathbf{1}[i_t\neq i_{t+1}]\leq m}}\sum_{t=1}^{T}
  \ell_t\left(\htheta_{i_t,t}\right) - \min_{\btheta \in \Theta}
  \sum_{t=1}^{T} \ell_t(\theta_t)}_{T_5} \label{Tracking_Decomp}
\end{align}
where the minimization in the second term of $T_4$ and first term of
$T_5$ is with respect to sequences of dynamical models with at most
$m$ switches, such that $\sum_{t=1}^T \mathbf{1}_{[i_t \neq
    i_{t+1}]}\leq m$.  In \eqref{Tracking_Decomp}, $T_4$ corresponds
to the tracking regret of our algorithm relative to the best sequence
of dynamical models within the DMD framework, and $T_5$ is the regret
of that sequence relative to the best comparator in the class
$\Theta_m$.
We use Corollary 3 of \cite{CesGaiLugSto12} to bound $T_4$, using $\eta_r=\sqrt{\frac{8\left((m+1)\log(N)+m\log(T)+1\right)}{T}}$ and $\lambda=m/(T-1)$. A slight
modification to their proof needs to be considered, because their
losses are bounded between $[0,1]$.  In our case, we assume our loss
function $\ell_t$ is convex, and Lipschitz on a bounded, closed, convex set
$\Theta$.  Therefore, we can say there exists a value $L$ such that $L
\triangleq \displaystyle\max_{t \in [1,T],\theta \in \Theta}
\ell_t(\theta) - \min_{t \in [1,T],\theta \in \Theta} \ell_t(\theta)$.
This value can be easily incorporated into the proofs and bounds of
\cite{CesGaiLugSto12} to give
{\allowdisplaybreaks
\begin{align}
T_4 \leq &L^2 \sqrt{\frac{T}{2} \left( (m+1) \log{N} + (T-1) h\left(\frac{m}{T-1}\right)
\right)}\nonumber\\
\leq &L^2\sqrt{\frac{T}{2} \left( (m+1) \log N + m \log T + 1\right)}
\nonumber
\end{align}}
where $h(x)=-x\log(x) - (1-x)\log(1-x)$ with respect to the natural logarithm.  $T_5$ can be bounded using Lemma~\ref{lem:main} on each time interval $[t_k,t_{k+1}-1]$ and
summing over the $m+1$ intervals, yielding
\begin{align}
T_5 \leq \frac{(m+1)\diam}{\eta_{T+1}} + \frac{2M}{\eta_T}V^{(m+1)}(\btheta_T) + \frac{G^2}{2\sigma} \sum_{t=1}^{T}
\eta_t. \qquad \Box
\nonumber
\end{align}

\subsection{Proof of Theorem~\ref{thm:covering}} 

Let $\alpha_*$ to be the dynamical parameter in our candidate 
covering set, $\mathcal{A}_N$ which minimizes the cumulative loss, $\alpha^*$ as the dynamical parameter in the entire space, $\mathcal{A}$
which minimizes total loss, and $\tilde{\alpha}$ to be the parameter in $\mathcal{A}_N$ closest to $\alpha^*$.  Formally, we use the following definitions:
\begin{align*}
\alpha_* \triangleq& \min_{\alpha \in \mathcal{A}_N} \sum_{t=1}^T \ell_t(\htheta_{\alpha,t}), \hspace{.1 in}
\alpha^* \triangleq \min_{\alpha \in \mathcal{A}} \sum_{t=1}^T\ell_t(\htheta_{\alpha,t}), \\
\tilde{\alpha}\triangleq&\min_{\alpha \in \mathcal{A}_N} \|\alpha-\alpha^*\|.
\end{align*} 
We decompose the regret in the following way:
\begin{subequations}
{\allowdisplaybreaks
\begin{align*}
&R_T(\THETA_T)=\underbrace{\displaystyle \sum_{t=1}^{T} \ell_t(\htheta_t) - \sum_{t=1}^{T} \ell_t\left(\htheta_{\alpha_*,t}\right)}_{T_6}\\
&+ \underbrace{\displaystyle\sum_{t=1}^{T} \ell_t\left(\htheta_{\alpha_*,t}\right) -
\sum_{t=1}^{T} \ell_t\left(\htheta_{\tilde{\alpha},t}\right)}_{T_7}
+\underbrace{\sum_{t=1}^T \ell_t\left(\htheta_{\tilde{\alpha},t}\right) - 
\sum_{t=1}^T \ell_t\left(\theta)\right)}_{T_8}
\end{align*}}
\end{subequations}

To bound $T_6$ we use the bounds of \cite[Corr 2.2]{CesLug06} for the
Exponentially Weighted Average Forecaster, which requires $\eta_r=\sqrt{\frac{2 \log N}{T}}$.  Similarly to the proof of
Theorem \ref{thm:dfs}, the bound needs to be adjusted to account for
the fact that our loss function, instead of being bounded by $[0,1]$
instead has arbitrary bounds $[\ell_{\min},\ell_{\max}]$.  Because,
$\ell_t$ is a convex function with a Lipschitz gradient defined on a bounded, closed, convex set, these
bounds are finite, and incorporating them into the proof of
\cite{CesLug06} is not difficult, yielding
\begin{align*}
T_6\leq (\ell_{\max} - \ell_{\max})\sqrt{2T \log(N)}.
\end{align*}
The term $T_7$ is upper bounded by $0$ by the definitions of $\alpha_*$ and $\tilde{\alpha}$.  Finally,
the bound on $T_8$ is just the DMD regret bound (Theorem \ref{thm:main}) with respect to $\tilde{\alpha}$:
\begin{align*}
T_8 \le  \frac{\diam}{\eta_{T+1}} + \frac{2M}{\eta_T} \sum_{t=1}^{T-1} \|\theta_{t+1} - \Phi_t(\theta_t,\tilde{\alpha})\| +
\frac{G^2}{2\sigma} \sum_{t=1}^{T} \eta_t.
\end{align*}

We now use the Lipschitz assumption on $\Phi_t$ to show the bound with respect to $\alpha^*$.  Notice the variation term can be separated.
\begin{align*}
V_{\Phi}(\btheta_T) =& \sum_{t=1}^{T-1} \|\Phi_t(\theta_t,\tilde{\alpha}) - \Phi_t(\theta_t,\alpha^*) + \Phi_t(\theta_t,\alpha^*) - \theta_{t+1}\|\\
\leq &\sum_{t=1}^{T-1} \|\Phi_t(\theta_t,\alpha^*)-\theta_{t+1}\| + T L \varepsilon_N
\end{align*} 
This shows we get a regret bound which scales like the variation from the best possible parameter $\alpha^*$, in the set $\mathcal{A}$.  Setting $\eta_t\propto\frac{1}{\sqrt{t}}$ gives the result.

\subsection{Proof of Lemma~\ref{lem:induction}} 
The proof is by induction, and we assume without loss of generality that $\beta = \zeros$. Assume that $\wh{\mu}_{\alpha,t} = \wh{\mu}_{0,t} + K_t \alpha$; this is trivially true for $t =1$ since $K_1 = \zeros$ and $\wh{\mu}_{\alpha,1} = \wh{\mu}_{\beta,1}$ by construction. If we use DMD with $\Phi_t(\theta,\alpha)$ for $t\ge1$, applying \eqref{eq:dmd} in this setting yields
{\allowdisplaybreaks
\begin{align*}
\wh{\mu}_{0,t+1} =& A_t[(1-\eta_t)\wh{\mu}_{0,t} + \eta_t \phi(x_t)] + c_t\\
\wh{\mu}_{\alpha,t+1} =& A_t[ (1-\eta_t)\wh{\mu}_{\alpha,t} + \eta_t \phi(x_t)] + B_t\alpha +c_t\\
 =& A_t[ (1-\eta_t)(\wh{\mu}_{0,t} +K_t\alpha) + \eta_t \phi(x_t)] + B_t\alpha +c_t\\
 =& A_t[ (1-\eta_t)\wh{\mu}_{0,t} + \eta_t \phi(x_t)]  +c_t  + (1-\eta_t) A_t K_t\alpha + B_t\alpha\\
 =& \wh{\mu}_{0,t+1} + (1-\eta_t) A_t K_t\alpha + B_t\alpha\\
 =& \wh{\mu}_{0,t+1} + K_{t+1}\alpha. \qquad \Box
 \end{align*}}
Notice that we must assume that $\htheta_{t}$ must lie
on the interior of the set $\Theta$ such that we can set the gradient
to 0 to find the minimizer of equation \ref{eq:dmd1} without
projecting back onto the set.

\subsection{Proof of Theorem~\ref{thm:additive}} 
Since $\wh{\mu}_{\alpha,t}$ is an affine function of $\alpha$ given
$\wh{\mu}_{\beta,t}$ for any $\beta$, and $\wt{\ell}_t(\mu)$ is a convex
function of $\mu$, we have that $g_t(\alpha)$ is a convex function of
$\alpha$.  For any $\alpha \in\reals^n$, we have
\begin{align*}
\sum_{t=1}^T &\ell_t(\wh{\theta}_t) - \sum_{t=1}^T \ell_t(\theta_t) \\
=& \underbrace{\sum_{t=1}^T \ell_t(\wh{\theta}_t) - \sum_{t=1}^T \ell_t(\wh{\theta}_{\alpha,t})}_{T_9} + \underbrace{\sum_{t=1}^T \ell_t(\wh{\theta}_{\alpha,t})-\sum_{t=1}^T \ell_t(\theta_t)}_{T_{10}}. \label{eq:additive}
\end{align*}
To bound $T_9$, note that $\wh{\theta}_t = \wh{\theta}_{\wh{\alpha}_t,t}$ and
$$\sum_{t=1}^T \ell_t(\wh{\theta}_t) - \sum_{t=1}^T \ell_t(\wh{\theta}_{\alpha,t}) = \sum_{t=1}^T \wt{\ell}_t(\wh{\mu}_t) - \sum_{t=1}^T \wt{\ell}_t(\wh{\mu}_{\alpha,t}).
$$
Furthermore, 
\begin{align*}
g_t(\alpha) =& \wt{\ell}_t(\wh{\mu}_{\alpha,t}) = \wt{\ell}_t(\wh{\mu}_{\wh{\alpha}_t,t}  + K_t (\alpha - \wh{\alpha}_t))
\end{align*}
 is convex in $\alpha$, where we have used Lemma~\ref{lem:induction}. Thus
$$T_9 \le \sum_{t=1}^T [\wt{\ell}_t(\wh{\mu}_t) -
\wt{\ell}_t(\wh{\mu}_{\alpha,t})] = \sum_{t=1}^T [g_t(\wh{\alpha_t}) - g_t(\alpha)] \le C_1 \sqrt{T}$$ 
via \cite{Zin03}.
The term $T_{10}$ can be bounded directly using
Theorem~\ref{thm:main}, yielding for any $\alpha$
$$T_{10} \le C_2 \sqrt{T} \left(1 + \sum_{t=1}^T \|\theta_{t+1} - \Phi_t(\theta_t,\alpha)\|\right). \qquad \Box$$

\bibliographystyle{ieeetr}
\bibliography{DynamicOCP,WillettRefs,PatchImaging,HawkesRefs}

\end{document}